\PassOptionsToPackage{table,dvipsnames}{xcolor}

\documentclass[11pt,a4paper]{article}
\usepackage{times,latexsym}
\usepackage{url}
\usepackage[T1]{fontenc}

\usepackage[acceptedWithA]{tacl2021v1}

\usepackage{xspace,mfirstuc,tabulary}

\usepackage{microtype}
\usepackage{hyperref}
\usepackage{url}
\usepackage{graphicx}
\usepackage{arydshln}
\usepackage{pifont}
\usepackage{adjustbox}
\usepackage{soul}
\usepackage{amsmath}
\usepackage{inconsolata}
\usepackage{changepage}
\usepackage[inline]{enumitem}
\usepackage{booktabs}
\usepackage{makecell}
\usepackage{colortbl}
\usepackage{mdframed}
\usepackage{soul}
\usepackage{tablefootnote}
\usepackage{tikz}
\usetikzlibrary{tikzmark}
\usepackage{array}
\usepackage{multirow}
\usepackage{float}
\usepackage{ifthen}

\usepackage{subcaption} 
\usepackage{graphicx}   

\newcommand{\showcomments}{1}
\newcommand\cm[1]{\ifthenelse{\showcomments=1}{\textcolor{magenta}{[CM: #1]}}{}}
\newcommand\kl[1]{\ifthenelse{\showcomments=1}{\textcolor{cyan}{[KL: #1]}}{}}
\newcommand\my[1]{\ifthenelse{\showcomments=1}{\textcolor{orange}{[MY: #1]}}{}}
\newcommand\mi[1]{\ifthenelse{\showcomments=1}{\textcolor{green}{[MI: #1]}}{}}
\newcommand\dr[1]{\ifthenelse{\showcomments=1}{\textcolor{purple}{[DR: #1]}}{}}
\newcommand\jc[1]{\ifthenelse{\showcomments=1}{\textcolor{red}{[JC: #1]}}{}}

\newif\iftaclinstructions
\taclinstructionsfalse 
\iftaclinstructions
\renewcommand{\confidential}{}
\renewcommand{\anonsubtext}{(No author info supplied here, for consistency with
TACL-submission anonymization requirements)}
\newcommand{\instr}
\fi

\iftaclpubformat 

\else

\fi

\def\sym#1{\ifmmode^{#1}\else\(^{#1}\)\fi}
\newcommand\contextemoji{\raisebox{-2pt}{\includegraphics[width=1em]{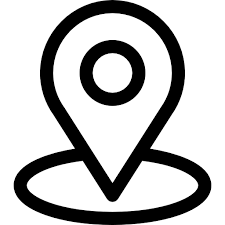}}}

\newcommand{\huggingface}{\raisebox{-1.5pt}{\includegraphics[height=1.05em]{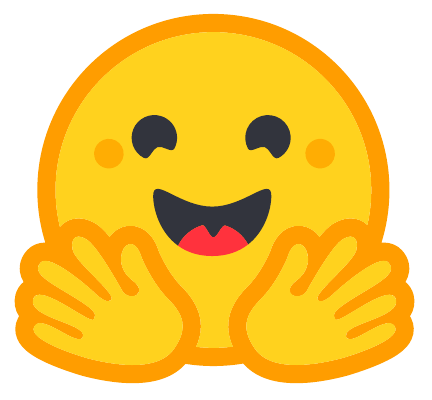}}\xspace}
\newcommand{\github}{\raisebox{-1.5pt}{\includegraphics[height=1.05em]{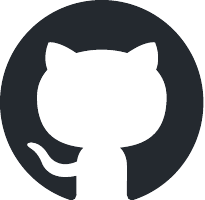}}\xspace}

\title{\contextemoji { Contextualized Evaluations: \\ Judging Language Model Responses to Underspecified Queries}}

\newcommand\pennemoji{\raisebox{-2pt}{\includegraphics[width=0.7em]{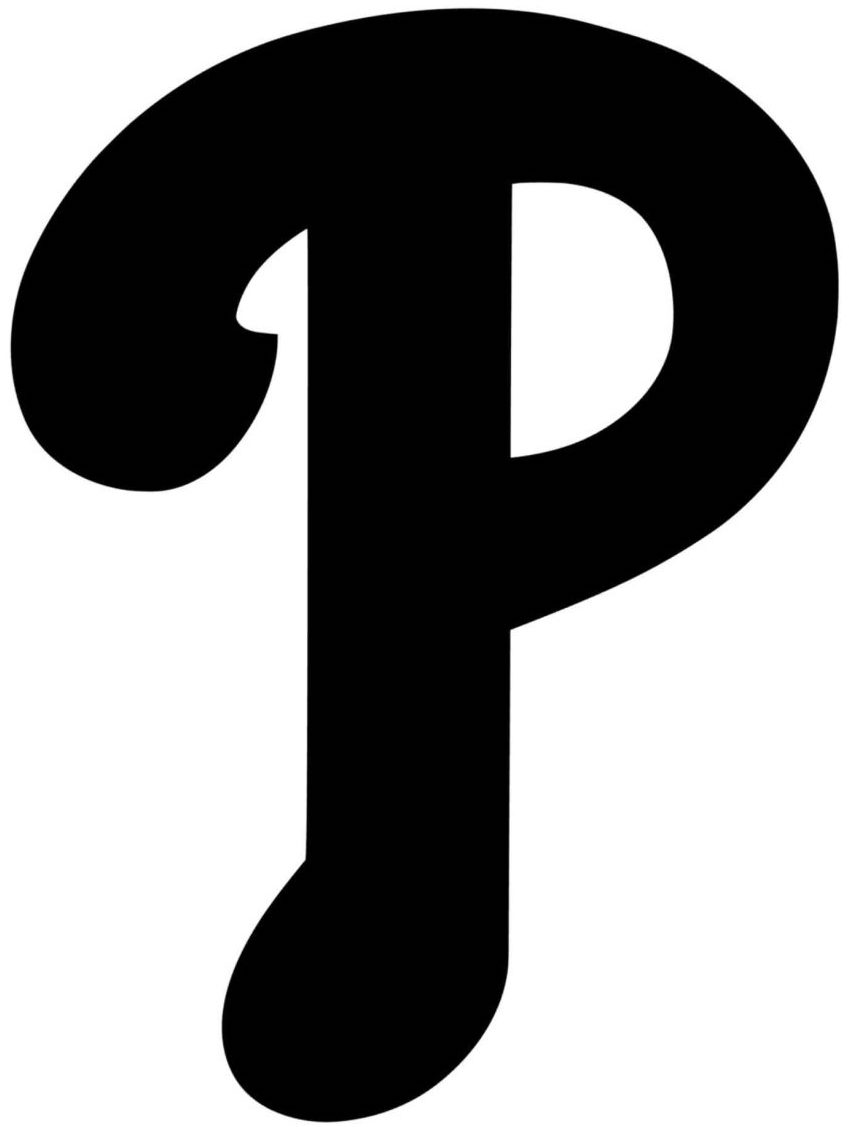}}}
\newcommand\aitwoemoji{\raisebox{-2pt}{\includegraphics[width=0.9em]{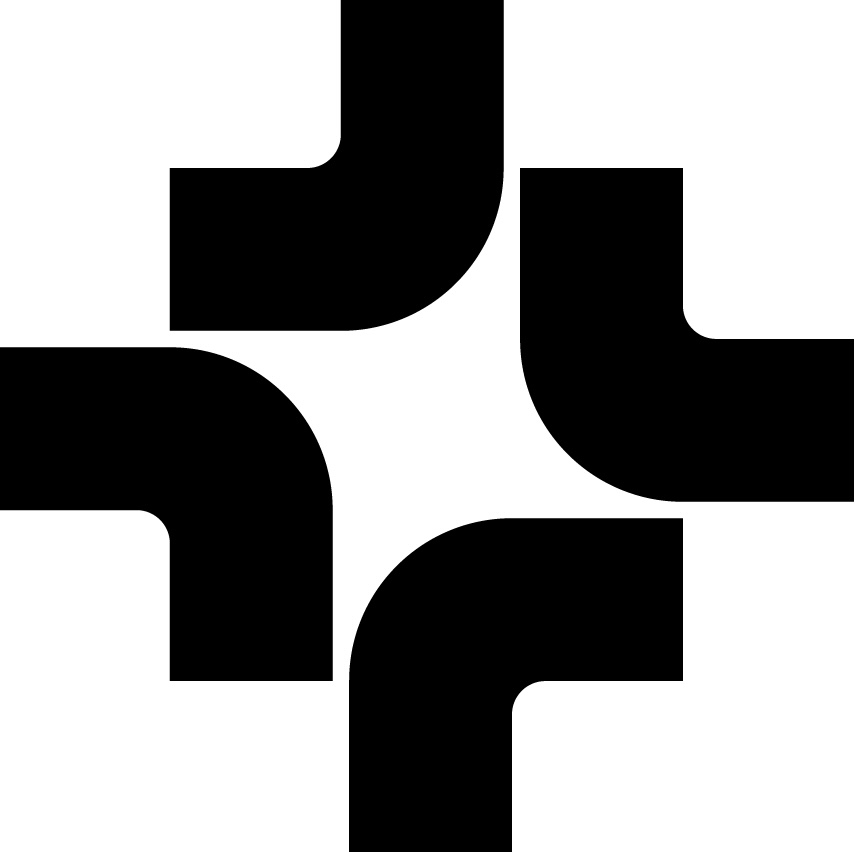}}}
\newcommand\umdemoji{\raisebox{-2pt}{\includegraphics[width=0.9em]{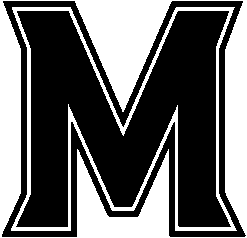}}}

\author{Chaitanya Malaviya\textsuperscript{\pennemoji}\thanks{\, Work done at the Allen Institute for AI.} \quad
    \textbf{Joseph Chee Chang}\textsuperscript{\aitwoemoji} \quad
    \textbf{Dan Roth}\textsuperscript{\pennemoji} \\
    \textbf{Mohit Iyyer}\textsuperscript{\umdemoji} \quad
    \textbf{Mark Yatskar}\textsuperscript{\pennemoji} 
    \quad
    \textbf{Kyle Lo}\textsuperscript{\aitwoemoji}
    \vspace{0.1in} \\
    \textsuperscript{\pennemoji}University of Pennsylvania \hspace{0.1in}
    \textsuperscript{\aitwoemoji}Allen Institute for AI \hspace{0.1in}
     \textsuperscript{\umdemoji}University of Maryland, College Park\\
    {\tt cmalaviy@seas.upenn.edu, kylel@allenai.org}\\
}

\begin{document}
\maketitle
\begin{abstract}
 Language model users often issue queries that lack specification, where the context under which a query was issued---such as the user's identity, the query's intent, and the criteria for a response to be useful---is not explicit.
  For instance, a good response to a subjective query like “\textit{What book should I read next?}” would depend on the user’s preferences, and a good response to an open-ended query like “\textit{How do antibiotics work against bacteria?}” would depend on the user's expertise. This makes evaluation of responses to such queries an ill-posed task, as evaluators may make arbitrary judgments about the response quality. 
  To remedy this, we present \textit{\textbf{contextualized evaluations}}, a protocol that synthetically constructs
  context surrounding an underspecified query and provides it during evaluation.
  We find that the presence of context can 
  1) alter conclusions drawn from evaluation, even flipping benchmark rankings between model pairs,
  2) nudge evaluators to make fewer judgments based on surface-level criteria, like style, and 3) provide new insights about model behavior across diverse contexts. 
  Specifically, our procedure suggests a potential bias towards WEIRD (Western, Educated, Industrialized, Rich and Democratic) contexts in models' ``default'' responses and we find that models are not equally sensitive to following different contexts, even when they are provided in prompts.\footnote{\,Our code and data are available at \url{https://www.github.com/allenai/ContextEval}.
  }

\begin{center}
\begin{tabular}{rl}
 \huggingface & \href{https://huggingface.co/datasets/allenai/ContextEval}{\path{allenai/ContextEval}} \\
 \github & \href{https://github.com/allenai/ContextEval}{\path{allenai/ContextEval}} \\
\end{tabular}
\end{center}

  
\end{abstract}

\section{Introduction}
\label{sec:intro}

\begin{figure}[t!]
    \centering
    \includegraphics[width=\columnwidth]{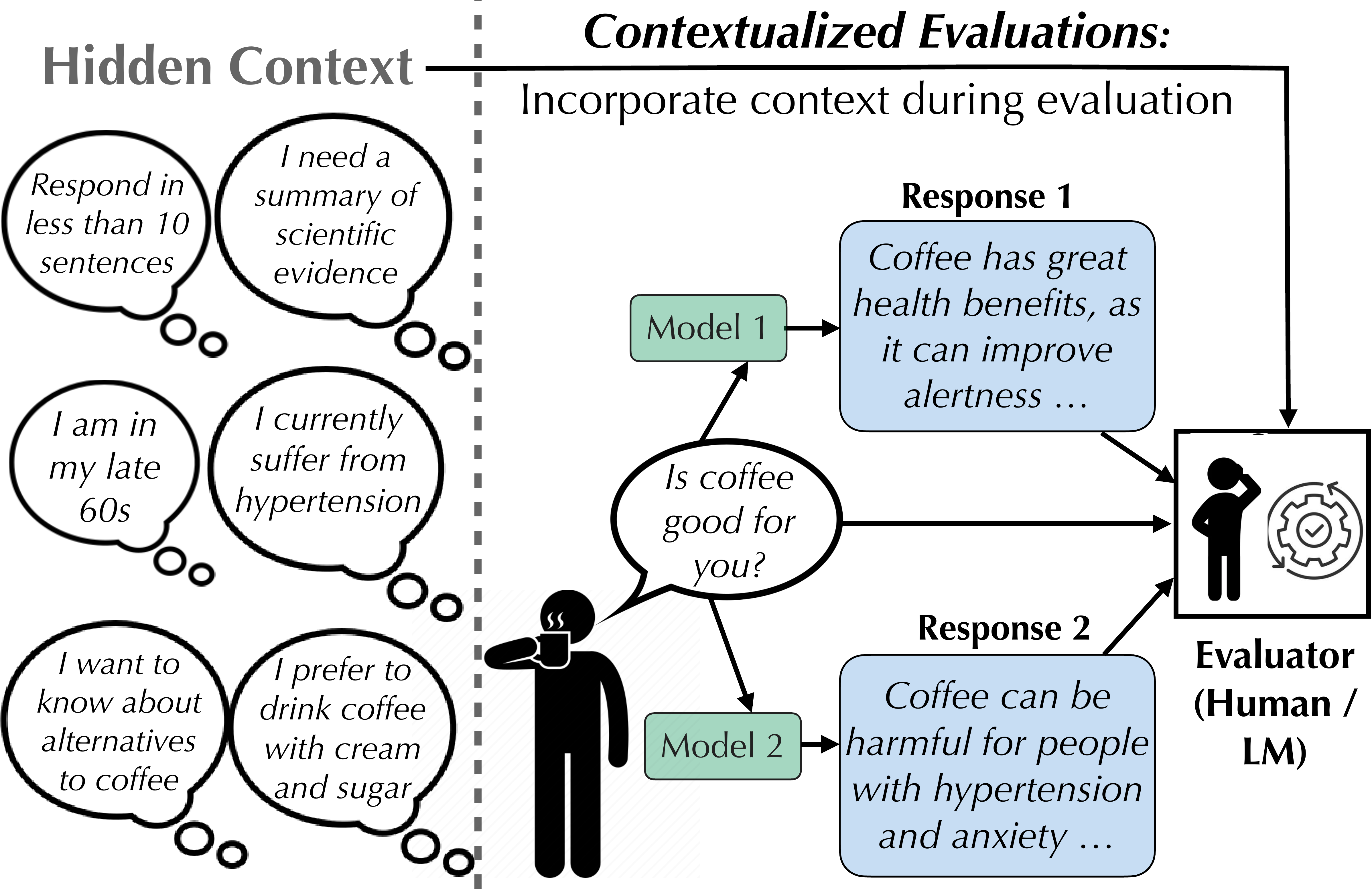}
    \caption{
    Queries issued by users to language models are often underspecified and can lead to arbitrary evaluation judgments of response quality.
    We present \textit{contextualized evaluations}, where queries are supplemented with surrounding context during evaluation.}
    \label{fig:problem}
\end{figure}

Users of language models often issue queries that are underspecified \cite{sparckjones07, clarke2009effectiveness, ziegler2019fine, keyvan2022approach,herlihy2024overcoming}, but common evaluation practices for language models do not account for this.
Consider an evaluator presented with a language model's response to an underspecified query such as ``\textit{Is coffee good for you?}'' (Figure~\ref{fig:problem}).
A language model might respond with an explanation about benefits like antioxidants and mental alertness, but this output would be unacceptable to users with certain health conditions.
Can an evaluator make an informed judgment about language model response quality without clear guidance on factors that reveal the intended user's preferences, background or other necessary criteria for the response to be useful?

In this work, we consider the role of \textbf{context} in the evaluation of language model responses to underspecified queries.
We propose \textbf{contextualized evaluations}---a protocol to synthetically generate and incorporate diverse contexts (represented as follow-up question-answer pairs) when evaluating model responses to underspecified queries.
By applying this procedure to queries from widely-used language model benchmark datasets, we investigate three main research questions:

First, we investigate whether providing context to evaluators has a substantial effect on the conclusions drawn from evaluation. We sample responses from pairs of language models and collect pairwise preference judgments from both model-based as well as human evaluators, in context-agnostic (only the query and model outputs) and context-aware settings (additionally with follow-up questions and answers to clarify the query).
Our experiments show not only that \textbf{inclusion of context during evaluation can significantly improve agreement between evaluators} (3-10\% absolute), but also that \textbf{context-aware evaluation can even flip benchmark rankings between model pairs} due to drastic changes in win rates.
This raises concern for the reliability of findings produced from context-agnostic evaluations using today's language model benchmarks, which we find to be full of underspecified queries (Table~\ref{tab:underspecification_types}).


Next, we ask whether context changes the criteria used by evaluators for making judgments. Without context, 
evaluators may make arbitrary judgments which reward surface-level adequacy of a response \cite{park-etal-2024-disentangling,chiang2024chatbot}. To evaluate this, we collect free-text justifications from both model and human evaluators. We find that \textbf{context-aware evaluation can decrease the frequency with which evaluators make judgments using surface-level properties like style} as opposed to other properties like response relevance (by 5-7\% absolute).


Finally, we investigate whether context helps us learn more about the ability of models to adapt to different user contexts. 
In our work, we use contexts to study biases exhibited by ``default'' model responses, i.e., those generated without context (\S\ref{sec:default}), to underspecified queries. We observe patterns suggesting a potential \textbf{bias where default model responses are better aligned with WEIRD (\textit{Western, Educated, Industrialized, Rich and Democratic}) contexts} \cite{henrich2010weirdest}, with weak to moderate agreement between human and model evaluators. Context can also allow us to directly evaluate the instruction-following and personalization abilities of models. We show this by evaluating sensitivity of models to different contextual attributes, where we find \textbf{considerable disparity in the ability of models to adapt to different contexts, even when they're provided as instructions in prompts} (\S\ref{sec:sensitivity}).



In summary, our findings suggest that underspecification can have a significant impact on the conclusions and insights drawn from evaluation. To address this, we propose \textbf{contextualized evaluations}, a simple and broadly applicable solution that involves synthesizing relevant context and injecting it into existing evaluation protocols (\S\ref{sec:context}). We show that context can increase agreement between evaluators and substantially modify model win rates based on pairwise preference judgements (\S\ref{subsec:results}). Further, we show that context can enable us to gather more insights about model behavior, such as identifying contexts that align more closely with default model responses (\S\ref{sec:default}) and assessing model sensitivity to different user contexts (\S\ref{sec:sensitivity}). Our work provides a plug-and-play recipe for incorporating context into future language model evaluations.



\section{How Prevalent is Underspecification?}
\label{sec:problem}

\setlength{\tabcolsep}{6pt} 

\renewcommand{\arraystretch}{1.15} 
\definecolor{Lavender}{rgb}{0.9,0.9,0.98}

\begin{table*}[t!]
\centering
\scalebox{.8}{
$\rotatebox[origin=c]{90}{\textit{Degree of underspecification}}%
\left\uparrow
\begin{tabular}{l@{~~}p{7cm}p{7cm}c}
\rowcolor{Lavender!55}
\multicolumn{1}{c}{\textbf{Query Type}} & \multicolumn{1}{c}{\textbf{Description}} & \multicolumn{1}{c}{\textbf{Example}} & \multicolumn{1}{c}{\textbf{Frequency}} \\ \toprule
 Incomplete & Missing essential information needed to provide a response (e.g., unresolved coreference). & \emph{What is the best team in the league?} & \cellcolor{Lavender!30}18.27\% \\
 Ambiguous & Can be interpreted in multiple ways (e.g., ambiguous word sense). & \emph{What is a transformer?} & \cellcolor{Lavender!5}1.87\% \\
 Subjective & Answers based on opinions or personal values (e.g., ``best'', ``worst''). & \emph{Who is the greatest philosopher from the 20th century?} & \cellcolor{Lavender!30}18.69\% \\
 Open-ended & Allows for multiple possible detailed responses (e.g., ``explain how'', ``describe why''). & \emph{Can you summarize recent work on mRNAs?} & \cellcolor{Lavender!70}76.17\% \\ \midrule
 Closed-ended & Requires specific, concise answer with little room for interpretation. & \emph{What is the capital of France?} & \cellcolor{Lavender!40}27.46\% \\
 \bottomrule
\end{tabular}
\right.\rotatebox[origin=c]{90}{}$
}
\caption{Queries from five LM benchmark datasets (Chatbot Arena, AlpacaEval, MTBench, ExpertQA and KIWI), categorized based on the amount of underspecification. 
Queries can present multiple types (e.g., ``\emph{What is the best team in the league?}'' is both incomplete and subjective).
}
\label{tab:underspecification_types}
\end{table*}

Users frequently issue underspecified queries, either to save time or because they are unsure of their information need. The nature of this underspecification can vary across queries, where at one extreme, some queries lack sufficient information to provide a meaningful response (e.g., ``best team in the league'') while at the other end, some queries are open-ended, allowing for many valid responses (e.g., ``summarize recent work on mRNAs''). 

We analyzed a total of 3580 queries by randomly sampling from five existing datasets used for benchmarking language models: Chatbot Arena \cite{chiang2024chatbot} (2500), AlpacaEval \cite{alpaca_eval} (645), MTBench \cite{zheng2023judging} (57), ExpertQA \cite{malaviya-etal-2024-expertqa} (300) and KIWI \cite{xu-etal-2024-kiwi} (78). 
We performed iterative qualitative coding to develop a schema over types of underspecification, and used \texttt{GPT-4o} to categorize queries (allowing for multiple label assignments).
Our results in Table~\ref{tab:underspecification_types} show that the \textbf{majority of queries in these benchmarks are open-ended}, while many are also incomplete and subjective.


Amid high prevalence of query underspecification, we posit three possible consequences to address in this work:

\begin{enumerate}[noitemsep, leftmargin=*]
    \item \textbf{Unreliable evaluation conclusions (\S\ref{subsec:results})}:
    Without context, 
    evaluators make subjective or arbitrary judgments that can result in
    inconsistent and unreliable model evaluations.
    \item \textbf{Evaluation focused on surface-level properties (\S\ref{subsec:results})}: 
    Lack of context results in evaluators making judgments of model responses based on surface-level criteria like style rather than whether user needs are fully met.
    \item \textbf{Limited assessment of contextual adaptability (\S\ref{sec:default}, \S\ref{sec:sensitivity})}: 
    When faced with underspecification, models default to generic responses, and as a result, their capacity to handle complex, diverse user preferences is not captured in context-agnostic evaluation.
\end{enumerate}


\section{Representing Context as Follow-Up QA}
\label{sec:context}


\setlength{\tabcolsep}{6pt} 
\renewcommand{\arraystretch}{1.15} 
\definecolor{Lavender}{rgb}{0.9,0.9,0.98}
\begin{table*}[t!]
\centering
\scalebox{0.75}{%
    \begin{tabular}{%
        p{0.15\textwidth}
        p{0.45\textwidth}
        p{0.15\textwidth}
        p{0.45\textwidth}
    }
    \multicolumn{2}{>{\cellcolor{Lavender!80}}c}{\textbf{Query Scope}} & \multicolumn{2}{>{\cellcolor{Lavender!80}}c}{\textbf{User Attributes}} \\
    \midrule
    \begin{tabular}[t]{@{}l@{}}\textbf{Focus \& Angle}\end{tabular} 
      & Would you also like information on the \emph{causes} of fever? 
      & \begin{tabular}[t]{@{}l@{}}\textbf{Expertise \&}\\\textbf{Familiarity}\end{tabular} 
      & What is your level of expertise in biology? \\
    \cellcolor{Lavender!80}\begin{tabular}[t]{@{}l@{}}\textbf{Style \& Tone}\end{tabular} 
      & \cellcolor{Lavender!80}Do you want this explanation to be technical or high-level? 
      & \cellcolor{Lavender!80}\begin{tabular}[t]{@{}l@{}}\textbf{Geographical}\\\textbf{Location}\end{tabular} 
      & \cellcolor{Lavender!80}Are you looking for global trends or trends in a specific region? \\
    \begin{tabular}[t]{@{}l@{}}\textbf{Intent}\end{tabular} 
      & Are you learning Flask for work or as a hobby? 
      & \begin{tabular}[t]{@{}l@{}}\textbf{Age Group}\end{tabular} 
      & What age group do you belong to? \\
    \cellcolor{Lavender!80}\begin{tabular}[t]{@{}l@{}}\textbf{Level of} \textbf{Detail}\end{tabular} 
      & \cellcolor{Lavender!80}How detailed do you want the response to be? 
      & \cellcolor{Lavender!80}\begin{tabular}[t]{@{}l@{}}\textbf{Language} \\ \textbf{Fluency}\end{tabular} 
      & \cellcolor{Lavender!80}Would you like the response in French? \\
    \begin{tabular}[t]{@{}l@{}}\textbf{Format}\end{tabular} 
      & Do you want a step-by-step overview or just a summary? 
      & \begin{tabular}[t]{@{}l@{}}\textbf{Interests}\end{tabular} 
      & What sports are you most interested in hearing about? \\
    \cellcolor{Lavender!80}\begin{tabular}[t]{@{}l@{}}\textbf{Need for}\\\textbf{Resources}\end{tabular} 
      & \cellcolor{Lavender!80}Would you like me to provide scientific studies and other references? 
      & \cellcolor{Lavender!80}\begin{tabular}[t]{@{}l@{}}\textbf{Cultural}\\\textbf{Background}\end{tabular} 
      & \cellcolor{Lavender!80}What cultural perspective should be considered in the response? \\
    \begin{tabular}[t]{@{}l@{}}\textbf{Clarification}\end{tabular} 
      & Do you mean apple the company or fruit? 
      & \begin{tabular}[t]{@{}l@{}}\textbf{Profession}\end{tabular} 
      & Are you asking as a researcher or developer? \\
    \cellcolor{Lavender!80}\begin{tabular}[t]{@{}l@{}}\textbf{Intended} \\ \textbf{Audience}\end{tabular} 
      & \cellcolor{Lavender!80}Are you preparing this for a general audience or experts? 
      & \cellcolor{Lavender!80}\begin{tabular}[t]{@{}l@{}}\textbf{Political} \textbf{Views}\end{tabular} 
      & \cellcolor{Lavender!80}Are you interested in the environmental perspective or economic impact? \\
    \begin{tabular}[t]{@{}l@{}}\textbf{Length}\end{tabular} 
      & How long would you like the summary to be? 
      & \begin{tabular}[t]{@{}l@{}}\textbf{Gender} \textbf{Identity}\end{tabular} 
      & Should I focus on health issues among people of a specific gender? \\
    \bottomrule
    \end{tabular}%
}
\caption{Taxonomy over types of common contextual attributes which can be lacking in underspecified queries, along with examples of follow-up questions for each attribute.}
\label{tab:taxonomy}
\end{table*}

\setlength{\tabcolsep}{6pt} 
\renewcommand{\arraystretch}{1.15} 
\definecolor{Lavender}{rgb}{0.9,0.9,0.98}

\begin{table*}[ht!]
\centering
\scalebox{0.75}{
\begin{tabular}{p{6cm} p{14cm}} 
\rowcolor{Lavender!80}
\multicolumn{1}{l}{\textbf{Query}} & \multicolumn{1}{l}{\textbf{Follow-Up QAs}} \\ \toprule
\multirow{3}{=}{\raggedright Give me a sample 5-day itinerary for a Switzerland holiday, starting from Basel.} 
& Q: What is your budget for the trip? \newline A: ["Economy", "Mid-range", "Luxury"] \\
& Q: What type of activities are you most interested in? \newline A: ["Outdoor activities", "Cultural experiences", "Historical sites", "Relaxation"] \\
& Q: Are you traveling alone or with others? \newline A: ["Alone", "With a partner", "With family", "With a group of friends"] \\ \hline

\rowcolor{Lavender!55}
\multirow{3}{=}{\raggedright I am going to make pumpkin pie for the first time. Can you help me? \newline (\texttt{AlpacaEval})}
& Q: Do you have any dietary restrictions or preferences? \newline A: ["None", "Gluten-free", "Dairy-free", "Vegan", "Low-sugar"] \\
\rowcolor{Lavender!55}
& Q: How many servings are you planning to make?\newline A: ["Small (4-6 servings)", "Medium (8-10 servings)", "Large (12+ servings)"] \\ 
\rowcolor{Lavender!55}
& Q: How much time do you have available for baking?\newline A: ["Under 1 hour", "1-2 hours", "More than 2 hours"] \\ \hline

\multirow{3}{=}{\raggedright How long will it take for a child to speak with therapy?} 
& Q: What is the initial diagnosis or reason for requiring therapy? \newline A: ["Speech Delay", "Autism Spectrum Disorder", "Hearing Impairment"] \\
& Q: How old is the child? \newline A: ["0-1 years", "2-3 years", "4-5 years", "6+ years"] \\
& Q: How frequently is the child receiving therapy? \newline A: ["Once a week", "Multiple times a week", "Occasionally", "Not receiving therapy yet"] \\
\bottomrule
\end{tabular}
}
\caption{Examples of follow-up QAs for a few underspecified queries. Each query has up to 10 such follow-up questions and answer sets associated with each question.}
\label{tab:context_examples}
\end{table*}



We represent context as follow-up question-answer (QA) pairs, simulating an interactive scenario where an agent can seek clarifying details from the user before responding.
For example, for a user issuing an underspecified query about coffee, we can represent relevant medical context:

\begin{quote}
User: ``\emph{Is coffee good for you?}''

$^*$Agent: ``\emph{Do you have hypertension?}''

$^*$User: ``\emph{Yes.}''
\end{quote}

\noindent where the follow-up QA pair (marked with $^*$) serves as additional context.

In practical scenarios, this context could be collected through user interaction or be stored by the agent as memory from past conversations. 



\subsection{Desiderata for Questions}

We require that each follow-up question is:
\begin{itemize}[noitemsep, leftmargin=*]
    \item \textbf{Salient}: The question must be relevant and important enough to the user query to warrant a response.
    \item \textbf{Actionable}: The question should anticipate that its answering will directly influence how the user's query best be addressed.
\end{itemize}

\noindent Context typically describes user attributes or the scope of the user's query (see Table~\ref{tab:taxonomy}). User attributes include the user's expertise, age group, location and other characteristics that affect the usefulness of the response. On the other hand, query scope describes the user's preferences for the response, such as the specific topic or aspect they want covered, the desired length or format, or the need for references. 


\subsection{Desiderata for Answers} 
While any one QA-pair constitutes a single contextual specification, when it comes to answers to follow-up questions, we impose requirements on the full \emph{answer set}:
\begin{itemize}[noitemsep, leftmargin=*, label=\small{$\bullet$}]
    \item \textbf{Realistic}: The answer choices should be plausible, such that a person could answer the question with any of the choices.
    \item \textbf{Complete}: The answer choices should cover a sufficient number of possible answers to the question.
    \item \textbf{Diverse}: The answer choices should be diverse, such that each answer would require adapting the response to the query in a different way.
\end{itemize}

\noindent Examples of follow-up questions and answer sets for a few queries are shown in Table~\ref{tab:context_examples}.

\subsection{Synthetically Generating Contexts}
\label{subsec:generating_context}

Scalably collecting many, diverse contexts for study can be difficult. We demonstrate and evaluate the use of language models in generating follow-up questions and their answer sets.

\paragraph{Approach} We perform few-shot prompting with \texttt{GPT-4o}, \texttt{Claude-3.5-Sonnet}, and \texttt{Gemini-1.5-Pro}); given a query, models are asked to first evaluate whether the query needs further context for generating a response, and if so, to generate a list of follow-up QAs. 
Our prompt is provided in the Appendix (Table~\ref{tab:contextual_followup_qa_prompt}).

\paragraph{Results} Out of the initial set of 3580 queries, we find that a total of 1881 queries need further context according to all three models. We use this set of 1881 queries for all further experiments. For these queries, we randomly sample a context from one of the models, which is then validated by a jury of models, which is also the same three models
in our case. These models provide a binary label for the importance of each follow-up question. We only retain those follow-up QAs which are found to be important by all three models.
There are an average of 9.32 follow-up QAs across queries.

\begin{figure}[t!]
    \centering
    \includegraphics[width=\columnwidth]{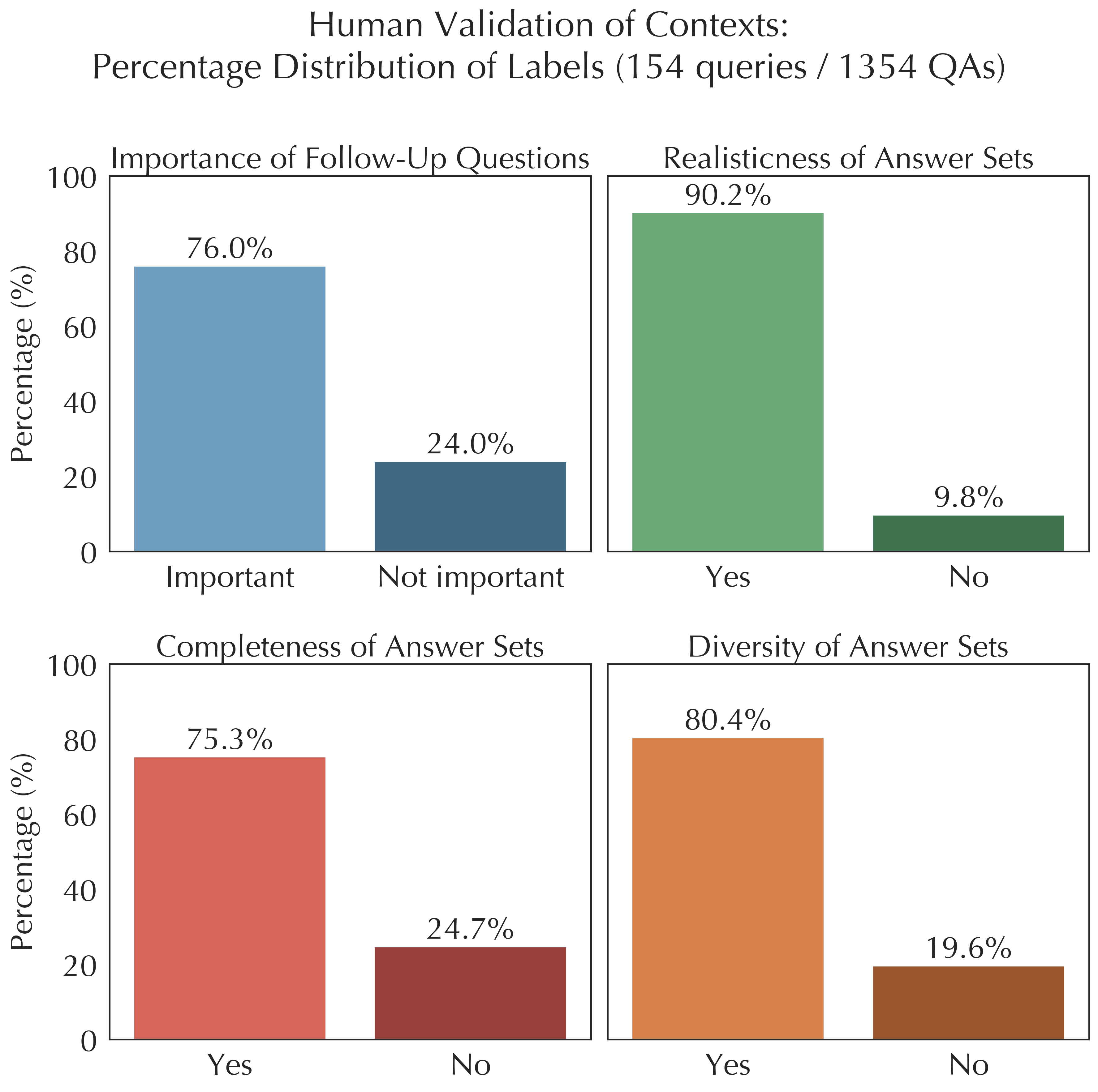}
    \caption{Human validation shows that most generated follow-up questions are important for clarifying an underspecified query and most generated answer sets are realistic, complete and diverse.}
    \label{fig:context_validation}
\end{figure}

\paragraph{Human evaluation}

To ensure that our contexts meet the criteria outlined in \S\ref{sec:context}, we ask 251 human annotators recruited from Prolific to validate contexts for a random sample of 154 underspecified queries, where each example is annotated by 3 annotators. 
Each annotator is shown the underspecified query and the corresponding follow-up QAs, and asked to provide binary labels for the importance of the follow-up question, and the realisticness, completeness and diversity of the answer sets. Further annotation details are provided in Appendix~\ref{app:annotation}. The majority label percentages based on this validation are shown in Figure~\ref{fig:context_validation}. We find that most follow-up questions ($\sim$76\%) are found to be important, while answer sets are mostly complete ($\sim$75\%), realistic ($\sim$90\%) and diverse ($\sim$80\%).
This gives us confidence to rely on these generated contexts for our studies. 

\section{Designing Contextualized Evaluations}
\label{sec:experiments}

\subsection{Evaluation Settings}
\label{subsec:settings}


\begin{figure}[ht!]
    \centering
    \includegraphics[width=0.8\columnwidth]{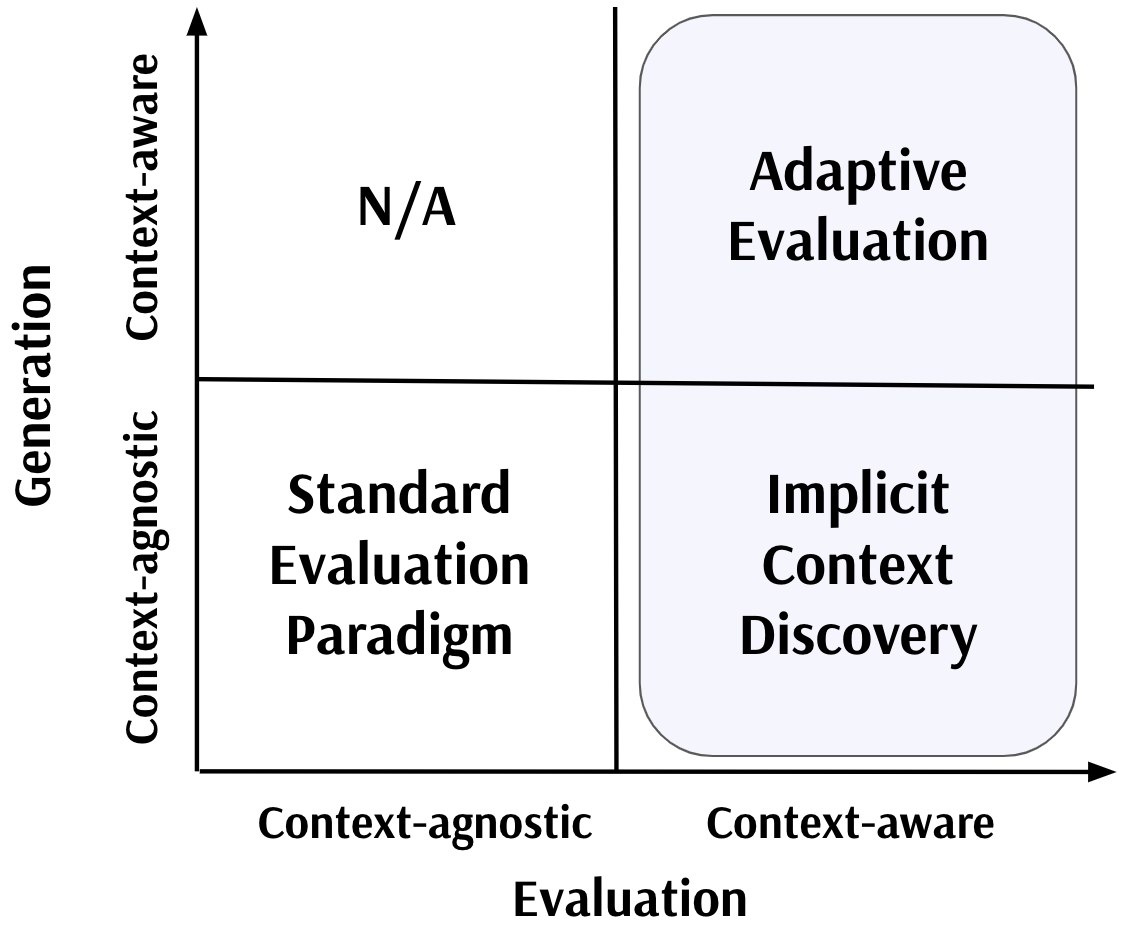}
    \caption{
    Our work defines two new evaluation settings---\emph{adaptive evaluation} and \emph{implicit context discovery}---distinctive from the \emph{standard evaluation paradigm} which is context-agnostic.
    }
    \label{fig:quadrants}
\end{figure}

Context can be provided during response generation, evaluation or both (illustrated in Figure~\ref{fig:quadrants}). Typical evaluation protocols are conducted in an entirely context-agnostic manner. We study two alternate evaluation protocols, \textit{adaptive evaluation}, where both generation and evaluation are context-aware and \textit{implicit context discovery}, where generation is context-agnostic but evaluators are provided context to rate default responses.

Context-agnostic responses are generated by prompting models to provide a response to the query alone, while context-aware responses are generated by providing a sampled context with each query. Specifically, we sample a \textit{single} answer to each follow-up question for a query using \texttt{GPT-4o}, such that the context is internally consistent, and provide this list of QA pairs to the model as context. Evaluators are required to perform pairwise judgments. During context-agnostic evaluation, evaluators are provided just the query and two responses, but additionally given the sampled context during context-aware evaluation. Prompts and experimental details are in Appendix~\ref{app:experimental}.

\begin{enumerate}[noitemsep, leftmargin=*]
    \item \raggedright \textbf{Standard Evaluation Paradigm} (\texttt{NoCtxGen-NoCtxEval}): 
    In a standard evaluation paradigm, responses to queries are generated without any supplemental context, and independent evaluators are not exposed to any such context either.
    \item \raggedright \textbf{Implicit Context Discovery} (\texttt{NoCtxGen-CtxEval}): In this setting, responses are generated without context but evaluators are exposed to plausible context surrounding the query. This context is fixed across both responses for pairwise evaluation. 
    Since models are left to assume contexts based on initial queries, we can use this setting to discover implicit biases in default model responses by providing context to the evaluators. We showcase an example of such an analysis in \S\ref{sec:default}.
    \item \raggedright \textbf{Adaptive Evaluation} (\texttt{CtxGen-CtxEval}): Finally, we consider a setting where both generative models and evaluators are exposed to context, which is fixed across model pairs. In this setting, models showcase how well they can adapt to user contexts, and hence, we can probe the instruction-following and personalization abilities of models.
\end{enumerate}

\subsection{Model Pairs to be Evaluated}
\label{subsec:response_gen}

We generate responses for all 1881 queries by zero-shot prompting three pairs of candidate models.
We selected model pairs based on their relative rankings (at the time of the study) on the Chatbot Arena Leaderboard:\footnote{\url{https://lmarena.ai/?leaderboard}}
\begin{itemize}[noitemsep, leftmargin=*]
    \item \texttt{GPT-4o} vs \texttt{Gemini-1.5-Flash}, which were Rank 3 vs Rank 4, selected as top-ranking proprietary models.
    \item \texttt{Claude-3.5-Sonnet} vs \texttt{Llama-3.1-405B}, which were Rank 5 vs Rank 6, selected as similarly ranked proprietary and open-weights model.
    \item \texttt{Gemma-2-27B} vs \texttt{Jamba-1.5-Large}, which were Rank 19 vs Rank 20, selected as similarly ranked open-weights contenders.
\end{itemize}


\subsection{Collecting Human Evaluator Judgments}
\label{sec:prolific-human-eval}

We recruit our human evaluators through Prolific and ask them to first complete a training task where they evaluate responses to queries in both context-agnostic and context-aware settings. In the context-agnostic setting, evaluators are asked to give an overall preference for one of the two responses or indicate a \textit{Tie}, along with a free-text justification. While in the context-aware setting, evaluators are first asked to evaluate whether the constraints in each one of the follow-up QAs are satisfied by each of the two responses. Once they have provided these ratings, they are required to similarly provide an overall preference along with a free-text justification. The order of the two responses is always randomized. Each evaluator is shown a single query and pairs of responses for one of the three evaluation settings, which is randomly chosen for the evaluator. Every example is judged by 3 different evaluators and we collect ratings for 100 examples for all 3 model pairs and 3 settings, for a total of 100 * 3 * 3 * 3 = 2700 ratings. Further human evaluation details and interface screenshots are in Appendix~\ref{app:annotation}.

\subsection{Collecting Model Evaluator Judgments}
\label{sec:autorater-model-eval}

The use of language model-based evaluators (also known as autoraters or LM-as-judge) 
is becoming increasingly prevalent. Hence, we also use them to judge responses in all three settings. We use 5 autoraters for each candidate model pair, which includes the 6 models listed in \S\ref{subsec:response_gen} and \texttt{Qwen2-72B-Instruct} \cite{yang2024qwen2}, but always excluding the two candidate models to avoid self-preference bias \cite{panickssery2024llm}. In all three evaluation settings, autoraters are instructed to give an overall preference for one of the two responses or indicate a \textit{Tie}, and then provide a free-text justification.




\definecolor{lightgreen}{RGB}{204, 255, 204} 
\definecolor{mediumgreen}{RGB}{102, 255, 102} 
\definecolor{darkgreen}{RGB}{0, 204, 0} 

\definecolor{lblue}{RGB}{204, 229, 255} 
\definecolor{mblue}{RGB}{102, 178, 255} 
\definecolor{dblue}{RGB}{0, 102, 204} 

\newcommand{\colorcellagreement}[1]{%
  \ifdim#1pt<65pt\cellcolor{lightgreen}#1\else%
  \ifdim#1pt<75pt\cellcolor{mediumgreen}#1\else%
  \cellcolor{darkgreen}#1\fi\fi}

\newcommand{\colorcellfleiss}[1]{%
  \ifdim#1pt<0.25pt\cellcolor{lblue}#1\else%
  \ifdim#1pt<0.4pt\cellcolor{mblue}#1\else%
  \cellcolor{dblue}#1\fi\fi}

\renewcommand{\arraystretch}{1.25} 

\begin{table*}[t]
    \centering
    
    \begin{subtable}[t]{1.0\linewidth}
        \centering
        \scalebox{.7}{
        \setlength\tabcolsep{3pt}
        \begin{tabular}{l *{3}{cc}}
            \toprule
            Evaluation Setting 
              & \makecell{\texttt{GPT-4o vs}\\\texttt{Gemini-1.5-Flash}} 
              & \makecell{$\Delta$}
              & \makecell{\texttt{Claude-3.5-Sonnet vs}\\\texttt{Llama-3.1-405B}}
              & \makecell{$\Delta$}
              & \makecell{\texttt{Gemma-2-27B vs}\\\texttt{Jamba-1.5-Large}}
              & \makecell{$\Delta$} \\
            \midrule
            \texttt{NoCtxGen-NoCtxEval}
              & 70.67 (73.33) & -
              & 65.99 (67.75) & -
              & 71.33 (73.95) & - \\
            \texttt{NoCtxGen-CtxEval}
              & 75.00 (77.15) & 4.33 (3.82)
              & 71.33 (72.84) & 5.34 (5.09)
              & 74.00 (75.48) & 2.67 (1.53) \\
            \texttt{CtxGen-CtxEval}
              & 76.00 (76.36) & 5.33 (3.03)
              & 74.67 (77.19) & 8.68* (9.44)*
              & 74.33 (75.79) & 3.00 (1.84) \\
            \bottomrule
        \end{tabular}
        }
        \caption{Human Evaluator}
    \end{subtable}
    
    \vspace{1em}
    
    \begin{subtable}[t]{1.0\linewidth}
        \centering
        \scalebox{.7}{
        \setlength\tabcolsep{3pt}
        \begin{tabular}{l *{3}{cc}}
            \toprule
            Evaluation Setting 
              & \makecell{\texttt{GPT-4o vs}\\\texttt{Gemini-1.5-Flash}} 
              & \makecell{$\Delta$}
              & \makecell{\texttt{Claude-3.5-Sonnet vs}\\\texttt{Llama-3.1-405B}}
              & \makecell{$\Delta$}
              & \makecell{\texttt{Gemma-2-27B vs}\\\texttt{Jamba-1.5-Large}}
              & \makecell{$\Delta$} \\
            \midrule
            \texttt{NoCtxGen-NoCtxEval}
              & 68.03 (68.73) & -
              & 64.00 (65.04) & -
              & 73.22 (74.40) & - \\
            \texttt{NoCtxGen-CtxEval}
              & 78.34 (78.40) & 10.31* (9.67)*
              & 74.57 (74.89) & 10.57* (9.85)*
              & 80.98 (81.01) & 7.76* (6.61)* \\
            \texttt{CtxGen-CtxEval}
              & 74.51 (74.56) & 6.48* (5.83)*
              & 74.93 (75.13) & 10.93* (10.09)*
              & 75.94 (75.99) & 2.72* (1.59)* \\
            \bottomrule
        \end{tabular}
        }
        \caption{Model Autorater}
    \end{subtable}

     \caption{Agreement rate (max 100\%) across context-agnostic and context-aware evaluation settings.
    Pairs of model output judged for overall preference by three human evaluators (\S\ref{sec:prolific-human-eval}) and five autoraters (\S\ref{sec:autorater-model-eval}); agreement calculated both excluding ties and including ties (in parentheses).
    $\Delta$ shows absolute increase in agreement rate over baseline context-agnostic setting (\texttt{NoCtxGen-NoCtxEval}); significant differences (p $<$ 0.05) indicated with *.}
    \label{tab:model_agreement}
\end{table*}

\renewcommand{\arraystretch}{1.2}


\begin{table*}[ht!]
    \centering
    \scalebox{.7}{
    \setlength\tabcolsep{3pt}
        \begin{tabular}{l *{3}{cc}}
            \toprule
            Evaluation Setting 
              & \makecell{\texttt{GPT-4o vs}\\\texttt{Gemini-1.5-Flash}} 
              & \makecell{$\Delta$}
              & \makecell{\texttt{Claude-3.5-Sonnet vs}\\\texttt{Llama-3.1-405B}}
              & \makecell{$\Delta$}
              & \makecell{\texttt{Gemma-2-27B vs}\\\texttt{Jamba-1.5-Large}}
              & \makecell{$\Delta$} \\
            \midrule
            \texttt{NoCtxGen-NoCtxEval}
              & 68.03 & -
              & 64.00 & -
              & 73.22 & - \\
            \texttt{NoCtxGen-CtxEval}
              & 82.42 & 14.39
              & 79.43 & 15.43
              & 82.05 & 8.83 \\
            \texttt{CtxGen-CtxEval}
              & 81.40 & 13.37
              & 77.33 & 13.33
              & 82.54 & 9.32 \\
            \bottomrule
        \end{tabular}
        }
    \caption{Agreement rates in human evaluator pairwise judgments (without ties) across context-aware evaluation settings, considering only those examples where the two responses differed by at least one in the number of follow-up QAs they satisfied. We find agreement is substantially higher for this subset of examples than the overall agreement rates reported in Table~\ref{tab:model_agreement}.}
    \label{tab:human_agreement_high_diff}
\end{table*}

\renewcommand{\arraystretch}{1.2}


\begin{table*}[t]
    \centering
    
    \begin{subtable}[t]{0.9\linewidth}
        \centering
        \scalebox{.75}{
        \setlength\tabcolsep{4pt}
        \begin{tabular}{l ccc ccc}
            \toprule
            Evaluation Setting 
              & \multicolumn{3}{c}{Human Evaluator} 
              & \multicolumn{3}{c}{Model Autorater} \\
            \cmidrule(lr){2-4} \cmidrule(lr){5-7}
            & \texttt{GPT-4o} & \texttt{Gemini-1.5-Flash} & Tie 
            & \texttt{GPT-4o} & \texttt{Gemini-1.5-Flash} & Tie \\
            \midrule
            \texttt{NoCtxGen-NoCtxEval} 
              & 40.24          & \textbf{48.78}              & 10.98 
              & 39.07          & \textbf{53.00}              & 7.92  \\
            \texttt{NoCtxGen-CtxEval}    
              & 40.86          & \textbf{50.54}                       & 8.60  
              & \textbf{53.69} & 46.04                       & 0.27  \\
            \texttt{CtxGen-CtxEval}      
              & \textbf{52.75} & 32.97                       & 14.29 
              & \textbf{68.05} & 31.79                       & 0.16  \\
            \bottomrule
        \end{tabular}
        }
    \end{subtable}
    
    \vspace{1em}
    
    \begin{subtable}[t]{0.9\linewidth}
        \centering
        \scalebox{.75}{
        \setlength\tabcolsep{4pt}
        \begin{tabular}{l ccc ccc}
            \toprule
            Evaluation Setting 
              & \multicolumn{3}{c}{Human Evaluator} 
              & \multicolumn{3}{c}{Model Autorater} \\
            \cmidrule(lr){2-4} \cmidrule(lr){5-7}
            & \texttt{Claude-3.5-Sonnet} & \texttt{Llama-3.1-405B} & Tie 
            & \texttt{Claude-3.5-Sonnet} & \texttt{Llama-3.1-405B} & Tie \\
            \midrule
            \texttt{NoCtxGen-NoCtxEval} 
              & \textbf{58.75}            & 37.50            & 3.75  
              & \textbf{47.36}   & 47.07            & 5.57  \\
            \texttt{NoCtxGen-CtxEval}    
              & \textbf{45.88}            & 36.47            & 17.65 
              & 44.86            & \textbf{54.57}   & 0.57  \\
            \texttt{CtxGen-CtxEval}      
              & \textbf{55.55}            & 24.44            & 20.00 
              & \textbf{75.29}   & 24.49            & 0.23  \\
            \bottomrule
        \end{tabular}
        }
    \end{subtable}
    
    \vspace{1em}
    
    \begin{subtable}[t]{0.9\linewidth}
        \centering
        \scalebox{.75}{
        \setlength\tabcolsep{4pt}
        \begin{tabular}{l ccc ccc}
            \toprule
            Evaluation Setting 
              & \multicolumn{3}{c}{Human Evaluator} 
              & \multicolumn{3}{c}{Model Autorater} \\
            \cmidrule(lr){2-4} \cmidrule(lr){5-7}
            & \texttt{Gemma-2-27B} & \texttt{Jamba-1.5-Large} & Tie 
            & \texttt{Gemma-2-27B} & \texttt{Jamba-1.5-Large} & Tie \\
            \midrule
            \texttt{NoCtxGen-NoCtxEval} 
              & 38.82             & \textbf{54.12}              & 7.06  
              & \textbf{55.16}    & 39.16              & 5.68  \\
            \texttt{NoCtxGen-CtxEval}    
              & 36.67             & \textbf{50.00}              & 13.33 
              & 46.65             & \textbf{52.91}     & 0.44  \\
            \texttt{CtxGen-CtxEval}      
              & \textbf{49.43}             & 34.48              & 16.09 
              & \textbf{63.89}    & 35.93              & 0.17  \\
            \bottomrule
        \end{tabular}
        }
    \end{subtable}
    
    \caption{Model win rates across context-agnostic and context-aware evaluation settings. 
    Pairs of model output judged for overall preference by three human evaluators (\S\ref{sec:prolific-human-eval}) and five autoraters (\S\ref{sec:autorater-model-eval}); wins are determined by majority vote.
    In each setting, the overall preferred model is indicated in \textbf{bold}.}
    \label{tab:model_win_rates}
\end{table*}

\section{How Does Context Change Evaluation Conclusions?}
\label{subsec:results}

\paragraph{Presence of context improves agreement between evaluators.} We report percentage agreement rates
for both autoraters and human evaluators in Table~\ref{tab:model_agreement}.\footnote{To calculate percentage agreement, we determine the majority label and then compute the percentage of judgments matching this label for each example. If all 3 human judgments choose the same option, this would be 100\% agreement; if 2 of 3 humans choose the same option but one differs, this would be 66\% agreement, and so on. Percentage agreement averages these scores across all examples.}
Significance testing is performed using a paired t-test and values significantly different than the \texttt{NoCtxGen-NoCtxEval} setting are indicated with a \sym{*} (p < 0.05). 
We find that context-aware evaluation can significantly increase the agreement between evaluators, especially LM-based autoraters and sometimes even human evaluators, suggesting that context is helpful in grounding evaluators together towards more consistent judgments. 

While both autoraters and human evaluators generally show increased agreement when provided with context, the magnitude of improvement can vary. For example, autoraters show an improvement of roughly 3–10 points in agreement once context is introduced, whereas humans exhibit a more modest increase of about 3–5 points. This discrepancy suggests that LM-based evaluators may be more sensitive to context than human evaluators, potentially because they systematically focus on whether explicit constraints in the context are met. 
To investigate this further, we compute human agreement considering only those examples where the two responses differed by at least one in the number of follow-up QAs they satisfied (reported in Table~\ref{tab:human_agreement_high_diff}). These agreement rates are much higher than the overall agreement rates in Table~\ref{tab:model_agreement}, suggesting that follow-up QAs can provide relatively more objective criteria to distinguish between responses.

\paragraph{Presence of context can change model win rates and flip benchmark rankings.}
Based on the majority votes from autoraters and human evaluators, we report win rates for all model pairs in Table~\ref{tab:model_win_rates}.
Note that we exclude those examples where there was no clear majority vote. Overall, the autorater and human evaluator win rates suggest that win rates can substantially vary across evaluation settings. Importantly, we find that relative rankings between pairs of models can flip in settings where evaluators are provided context versus when they are not. For instance, in Table~\ref{tab:model_win_rates}, \texttt{GPT-4o} has a lower win rate than \texttt{Gemini-1.5-Flash} in the setting \texttt{NoCtxGen-NoCtxEval} but has a much higher win rate in the setting \texttt{CtxGen-CtxEval}. 

We observe similar trends between human evaluators and autoraters when examining magnitude of win rate shifts across evaluation settings, but note some key differences. 
For instance, while autoraters show a more significant flip favoring \texttt{GPT-4o} once context is provided, human evaluation shows a milder change, and is accompanied by an increased tie rate. When both responses incorporate the provided context sufficiently well, humans are less inclined to form a clear preference, resulting in a higher tie rate. On the other hand, autoraters rely on explicit checks for context fulfillment, which accentuates small differences in how each response incorporates context. 

\begin{figure*}[t!]
    \centering
    \includegraphics[width=0.5\paperwidth]{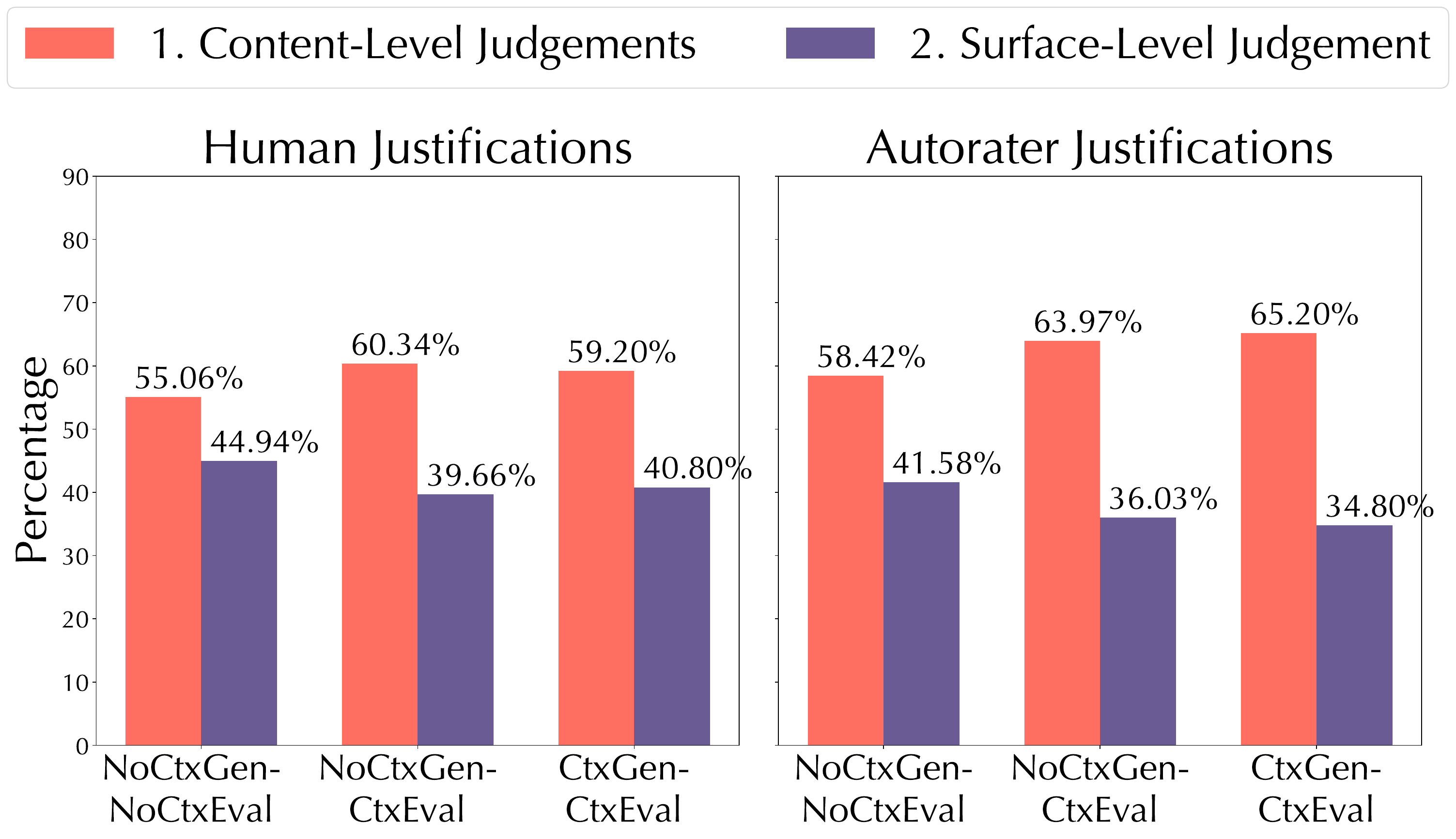}
    \caption{Types of human and autorater justifications across all evaluations settings. Note that there is a lower percentage of justifications based on surface-level criteria for the context-aware evaluation settings.}
    \label{fig:justifications}
\end{figure*}

\begin{figure*}[t!]
    \centering
    \includegraphics[width=2\columnwidth]{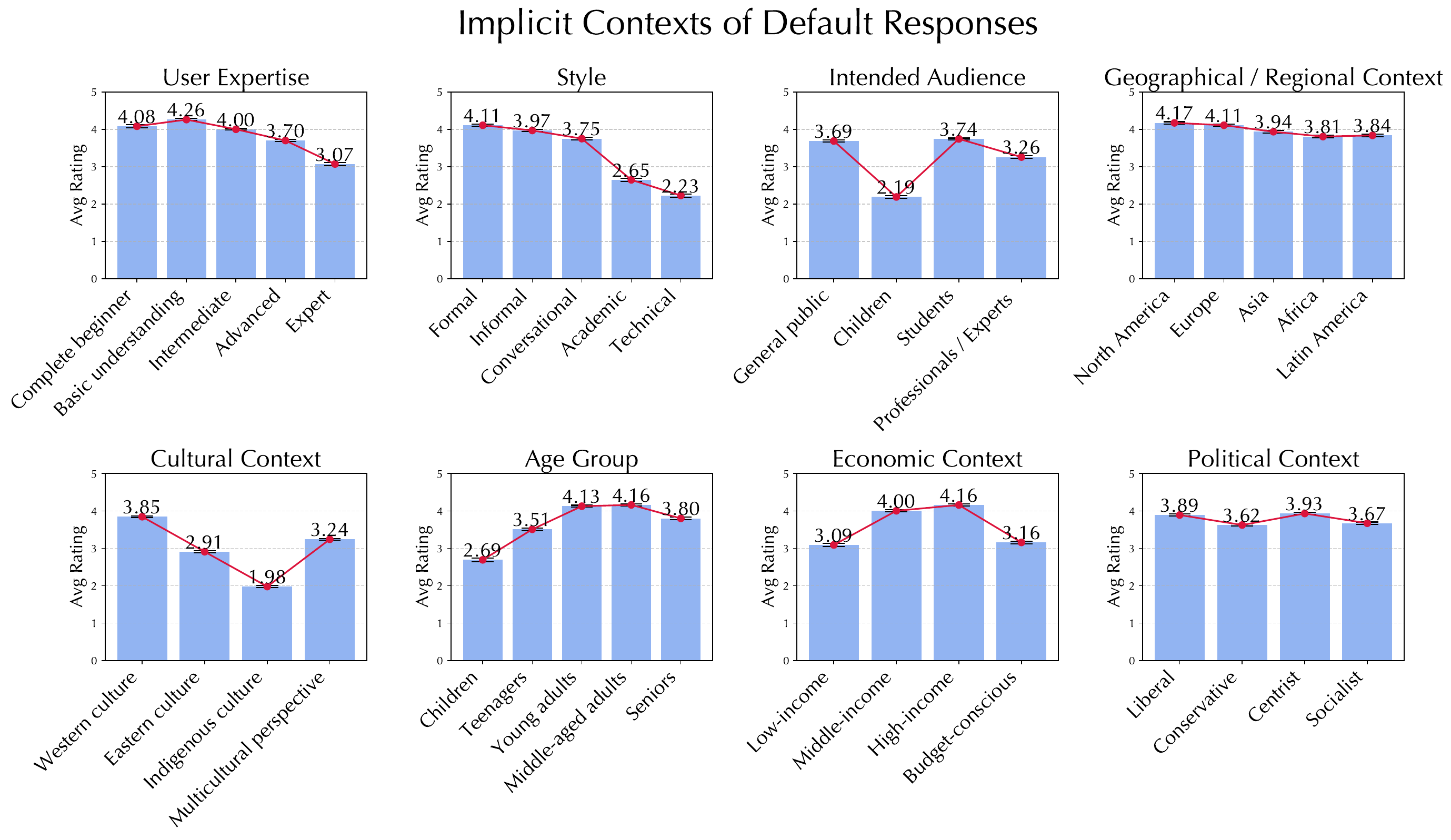}
    \caption{
    Relevance ratings of default responses from \texttt{GPT-4o} across various contextual attributes, as rated by \texttt{Gemini-1.5-Pro}. These plots suggest that default \texttt{GPT-4o} responses are better aligned towards users from western cultural contexts, high-income individuals and young and middle-aged adults.}
    \label{fig:default_contexts}
\end{figure*}

\paragraph{Presence of context can help evaluators avoid reliance on surface-level criteria to judge responses.} Next, we investigate whether context changes the criteria used by evaluators to make judgments. To do this, we analyze the free-text justifications provided by human evaluators as well as autoraters. We automatically code these justifications into two categories: 1) \emph{surface-level criteria} which includes criteria such as clarity, conciseness, style or formatting, tone and length, and 2) \emph{content-level criteria}, which includes criteria such as relevance, correctness, completeness, level of detail and context adherence. We then automatically classify justifications into these two categories using \texttt{GPT-4o}. Trends across all evaluation settings for both human evaluators and autoraters are shown in Figure~\ref{fig:justifications}. We note that in context-aware evaluation settings, there is a lower percentage of justifications that are based on surface-level criteria and more that are based on content-level criteria.
\section{Which Contexts are ``Default''?}
\label{sec:default}

Prior work has proposed methods to identify biases in language models through prompts tailored to uncover biases \cite{li-etal-2020-unqovering,santurkar2023whose,deshpande-etal-2023-toxicity,cheng2023everyone,durmus2024towards}. Instead, we study what contexts do models default to when presented with underspecified queries. We investigate the prevalence of these implicit contexts in default model responses using the evaluation setting \texttt{NoCtxGen-CtxEval} where responses are generated without context, but evaluators are shown context specific to a contextual attribute (such as "User Expertise").

\subsection{Methodology}
To investigate these implicit biases, we first define a list of contextual attributes and corresponding follow-up QAs (listed in Table~\ref{tab:default_context_qa} and based on Table~\ref{tab:taxonomy}). For each attribute, we first filter for those queries where this attribute is important to respond to the query and where the query is independent of the answer choices. For instance, a query such as "\textit{What is distillation in machine learning?}" would correspond to the attribute "User Expertise". Filtering is done using \texttt{GPT-4o} on 23,935 queries from all 5 datasets used in earlier experiments (prompt in Table~\ref{tab:query_followup_evaluation_prompt}). We then sample up to 1000 filtered queries randomly for each attribute and generate default responses (without context) for these queries using a candidate model.

To evaluate these default responses, we ask an automatic evaluator to provide absolute ratings for the default response (on a scale of 1-5) for response relevance for every value of each contextual attribute (prompt in Table~\ref{tab:response_quality_rating_prompt}). For instance, for a contextual attribute like "User Expertise", the evaluator would be asked to provide a rating for every possible value of this attribute ("Complete beginner", "Basic Understanding", ... , "Expert"). We then compute the average rating for every value of each contextual attribute and plot these trends in Figure~\ref{fig:default_contexts}. These results use \texttt{GPT-4o} as the candidate model, and \texttt{Gemini-1.5-Pro} as the evaluator.

For a subset of three contextual attributes ("Age Group", "Economic Context", "Cultural Context"), we also collect Likert ratings from human evaluators for the relevance of default responses to different possible values of each attribute (for e.g., possible values for "Age Group" -> [Children, Teenagers, Young adults, Middle-aged adults, Seniors]) for 75 queries (25 per contextual attribute). We then compute Spearman’s $p$ between human ratings for each query with the autorater judgements on the same query. Across the three contextual attributes, we obtain weak to moderate correlations (Age Group = 0.651, Cultural Context = 0.531, Economic Context = 0.276). We would like to emphasize that this task is highly subjective, which might affect the extent to which human ratings correlate with autorater ratings. At the same time, we find it encouraging that a subset of these attributes do have moderately positive correlations, which gives more credence to our findings below.

\subsection{Findings}
First, we note that default responses are better catered to users who have a basic understanding of the topic of a query as opposed to experts, and the language they use lacks technical depth. Further, our results suggest potential presence of a WEIRD-like bias, where default responses are better aligned with western cultural contexts, middle-to-high income individuals, and young and middle-aged adults. However, stronger correlations between human and model evaluators are required to strengthen this conclusion. Since context-agnostic evaluations might overlook disparities in how well default responses serve different contexts, we recommend future work to conduct similar analysis to discover implicit biases in default model responses.
\section{Which Contexts are Harder to Follow?}
\label{sec:sensitivity}

\subsection{Methodology}

A common approach to adapting models to different user contexts is providing these contexts as part of a prompt;
models that are capable of adapting to various user contexts are broadly more useful for more individuals. 
Context-agnostic evaluations can miss out on capturing the adaptability of models to various contexts. We use our evaluation setting \texttt{CtxGen-CtxEval} to investigate how robustly models can adapt to different values of a contextual attribute. Similar to the analysis in \S\ref{sec:default}, we use the contextual attributes and follow-up QAs defined in Table~\ref{tab:default_context_qa}. However, we now generate adapted responses for every value of each contextual attribute where the follow-up QA (e.g., ``\emph{What is your level of expertise on this topic?}'') and a specific answer (e.g., ``\emph{Complete beginner}'') is provided to the generator.

\begin{figure*}[t!]
    \centering
    \includegraphics[width=0.5\paperwidth]{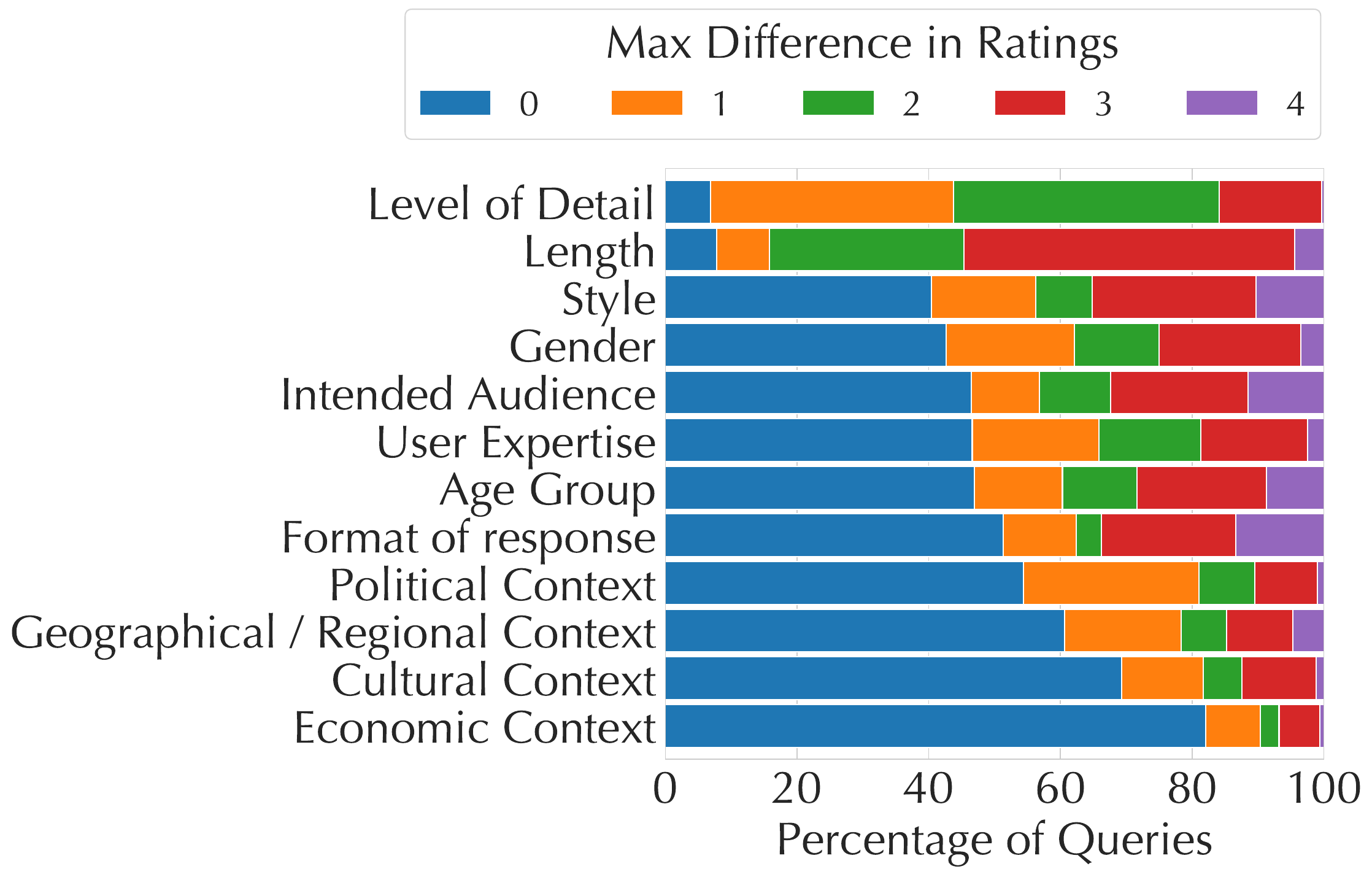}
    \caption{Distribution of the maximum difference in ratings between all values for each contextual attribute. 
    Contextual attributes that are more \textbf{difficult} for models to adapt to diverse values will have high percentage of Red (3) and Purple (4), while contextual attributes that are \textbf{easy} for models to adapt to will exhibit high percentage of Blue (0).
    We find that several contextual attributes show considerable disparity in models' ability to adapt their responses to different values of a contextual attribute.}
    \label{fig:variance}
\end{figure*}

We similarly ask an automatic evaluator to provide absolute ratings for an adapted response (on a scale of 1-5) for response relevance for the corresponding value of each contextual attribute. 
That is, in our example above, we would obtain the rating for if the contextual attribute value were ``\emph{Complete beginner}'', ``\emph{Intermediate}'', ``\emph{Expert}'', and so on, resulting in many ratings per follow-up QA.
We then compute the maximum difference of ratings across all values of a contextual attribute and plot these differences in Figure~\ref{fig:variance}. 
Larger differences indicate worse ability to adapt equally to all possible values of a contextual attribute. 
For example, for the follow-up question above, if the ratings to model responses were all ``4'', then the difference would be zero, meaning a model is equally adept at adapting to all expertise levels.
But if the responses were ``5'', ``5'', ``3'', ``3'', and ``2'', then the difference would be three, which indicates existence of a failure to adapt equally well to possible values of this contextual attribute.
These results also use \texttt{GPT-4o} as the candidate model, and \texttt{Gemini-1.5-Pro} as the evaluator.


\subsection{Findings}

First, we note that there is considerable disparity in the ability of \texttt{GPT-4o} to cater to different values of a contextual attribute. This suggests that the model is not equally good at adapting to all possible values of a contextual attribute. This disparity is larger for attributes such as length and level of detail, as well as gender and age group. The higher disparity in length and detail suggests that the model is not consistently able to adjust its responses based on the level of depth or brevity required by the user, often delivering overly simplistic or verbose responses. On the other hand, the differential performance of the model for attributes such as gender and age group can lead to unequal user experiences. Going forward, we suggest that future model evaluations conduct granular analysis of context adaptability to better capture the full spectrum of a model's capabilities.
\section{Related Work}
\label{sec:related}

\paragraph{Context as clarification questions.} A substantial amount of prior work has proposed methods to automatically generate clarification questions for tasks such as question answering \cite{rao-daume-iii-2018-learning,yu-etal-2020-interactive,kumar-black-2020-clarq,white-etal-2021-open,zhang2023clarify,zhang2024modeling,andukuri2024stargate}, information retrieval \cite{zamani2020generating,chi2024clarinet}, open-domain dialog generation \cite{aliannejadi2019asking,testoni-fernandez-2024-asking} and moral reasoning \cite{pyatkin-etal-2023-clarifydelphi}. Much of this prior work has focused on generating questions about ambiguous or incomplete inputs \cite{majumder-etal-2021-ask,zhang2023clarify}. While we also represent contexts as question-answer pairs, the intent of our questions extends beyond resolution of ambiguity, where they often just seek further exposition about the user's query.

\paragraph{Instruction-following and personalization.} Our study of context and its role in evaluation is relevant to prior work evaluating instruction-following and personalization abilities of models. The constraints represented through our context can be perceived as writing instructions or user attributes. Evaluations such as \citet{zhou2023instruction,suri24} have been useful for measuring instruction-following abilities of models. The contexts considered in our work differ from instructions in existing work, as our context is intended to increase the amount of specification for underspecified queries for better evaluation, rather than probing model abilities to follow writing instructions.

Work on personalization has advocated for evaluating the adaptability of model responses to users' personalized contexts \cite{flek-2020-returning,salemi-etal-2024-lamp} and proposed training methods for better alignment with user contexts \cite{lee2022divdis,cheng2023everyone,siththaranjan2023dpl,jang2023personalized,hwang-etal-2023-aligning,pitis2024improving}. We concur that evaluating adaptability of models to diverse contexts is important and our work presents a framework to conduct such an evaluation.

\paragraph{LM-based Autoraters.} Language model-based autoraters (also called LM-as-judge) hold great promise in improving the efficiency and reducing the cost of evaluations \cite{chiang-lee-2023-large}. Prior work has identified that these autoraters are subject to biases such as preferring longer responses \cite{dubois2024lengthcontrolled}, preferring their own responses \cite{panickssery2024llm} and others \cite{zheng2023judging,shen-etal-2023-large,wang-etal-2024-large-language-models-fair}. Methods have been proposed to overcome some of these biases by providing clearer criteria during evaluation \cite{liu-etal-2023-g,saha-etal-2024-branch}. Our work proposes a simple recipe for improving the reliability of LM-based autoraters by providing context surrounding queries.

\section{Limitations}
\label{sec:limitations}

\paragraph{Scope of Context.} The contextual attributes considered in our work only represent a sample of common information types that are missing in queries, which can be helpful to respond to a user’s query. The taxonomy presented in Table~\ref{tab:taxonomy} is not meant to be exhaustive and there are likely other types of context that are important.

\paragraph{Effect of Amount of Context.} In our work, we only consider up to 10 follow-up QAs for each query. We did not analyze the effect of the number of follow-up QAs on the results presented in section~\ref{sec:experiments}. In future work, it would be worthwhile to analyze how our results vary with the amount of specification in the context.

\paragraph{Autorater Reliance.} Language model-based autoraters improve the efficiency and decrease the cost of evaluation. Hence, we relied primarily on these autoraters for a subset of our analysis (presented in \S\ref{sec:default} and \S\ref{sec:sensitivity}). Human evaluation will strengthen the conclusions made in these analyses, and we will consider this in future work.

\section{Conclusion}
\label{sec:conclusion}

Through our study, we showed that existing query datasets contain many underspecified queries, and this underspecification can have significant impacts on the nature of our evaluation. We present \textbf{contextualized evaluations}, a simple, plug-and-play recipe to enrich queries with context in the form of follow-up question-answer pairs. Our experiments show that context can change the conclusions we draw from evaluations, and decrease the extent to which evaluators rely on surface-level criteria to judge responses. Finally, we show that context can help identify implicit biases in default model responses. 
We suggest future work to perform contextualized evaluations for a holistic understanding of how well models adapt to diverse user contexts.

\section{Acknowledgments}
First, we thank all the human annotators who took time to participate in our studies. We also thank the following people for helpful discussions and comments: Sihao Chen, Doug Downey, Greg Durrett, Varsha Kishore, Nelson Liu, Nicholas Lourie, Vishakh Padmakumar, Valentina Pyatkin, Elizabeth Sieber, Luca Soldaini, David Wadden, Dan Weld, Michael Zhang and members of the Semantic Scholar research group at Ai2. Finally, we are grateful to the TACL reviewers and our action editor Omri Abend for their thoughtful comments and suggestions.

\bibliography{custom}
\bibliographystyle{acl_natbib}

\clearpage

\appendix

\section{Human Evaluation Details}
\label{app:annotation}

\paragraph{Participants.} We recruited a total of 1,085 participants through the Prolific crowdsourcing platform for the human evaluation studies. Evaluators were required to have an approval rate of 99\% and 500 prior successful submissions on the platform. They were also required to be fluent in English. Most participants came from the US and UK. 

\paragraph{Evaluation Setup.} For the main preference judgment task, we sampled a query and evaluation setting (from the three settings discussed in \ref{sec:experiments}) randomly for each evaluator. Evaluators were also given a choice to skip a query for a different one, if they were unfamiliar with the topic of the original query. Each query-response pair was judged by 3 different evaluators and a single evaluator was allowed to complete up to 3 examples. Before starting the task, evaluators were required to complete a training task in the context-agnostic and context-aware evaluation settings, where they were given feedback about their ratings. Evaluators were paid \$1.75 for each example at a rate of \$15 per hour, where they were allocated a total of 7 minutes per example.

\begin{figure*}[ht!]
    \centering
    \includegraphics[scale=0.45]{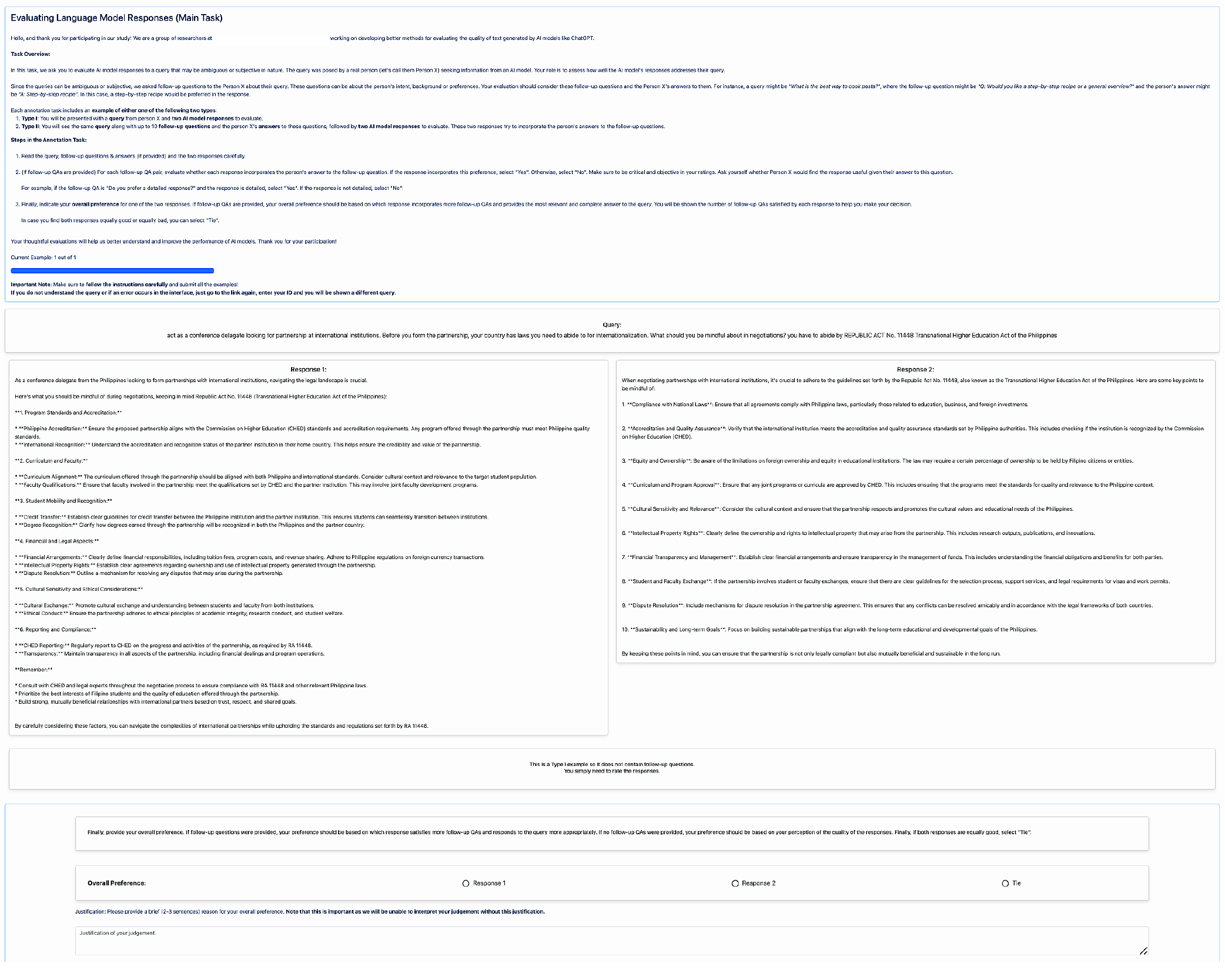}
    \caption{Screenshot of the evaluation interface used for context-agnostic evaluation.}
    \label{fig:screenshot_wo_ctx}
\end{figure*}

\begin{figure*}[ht!]
    \centering
    \includegraphics[scale=0.2]
    {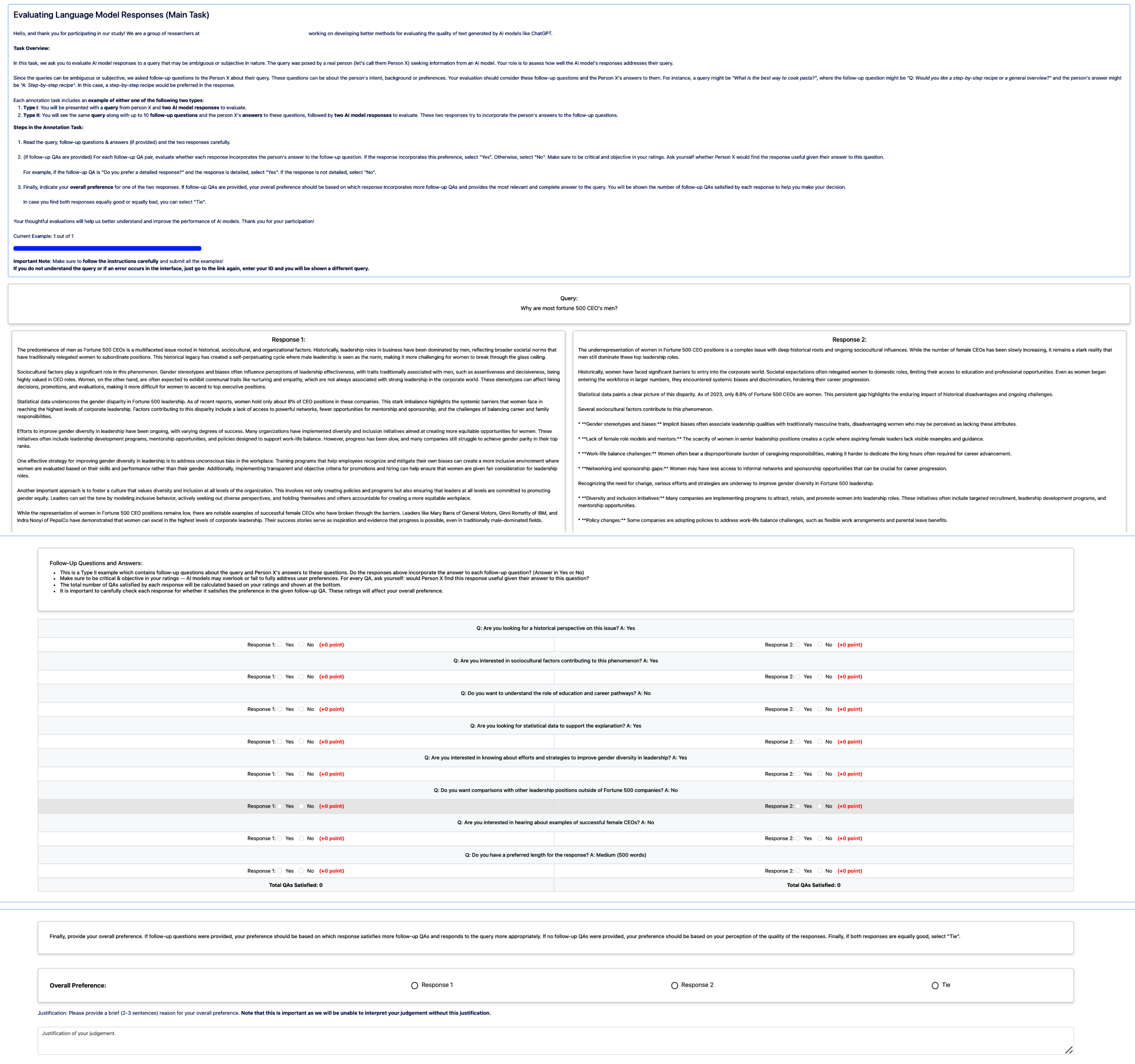}
    \caption{Screenshot of the evaluation interface used for context-aware evaluation.}
    \label{fig:screenshot_w_ctx}
\end{figure*}

\paragraph{Interface.} Screenshots of the evaluation interfaces for context-agnostic and context-aware evaluation are provided in Figures~\ref{fig:screenshot_wo_ctx} and \ref{fig:screenshot_w_ctx} respectively.

\section{Experimental Details}
\label{app:experimental}

\paragraph{Models.} We used 3 pairs of models for pairwise evaluations: \texttt{GPT-4o} \cite{achiam2023gpt} and \texttt{Gemini-1.5-Flash} \cite{reid2024gemini}, \texttt{Claude-3.5-Sonnet} \cite{TheC3} and \texttt{Llama-3.1-405B} \cite{dubey2024llama}, \texttt{Gemma-2-27B} \cite{team2024gemma} and \texttt{Jamba-1.5-Large} \cite{team2024jamba}.  We list the identifiers of the models that were used for experiments in Table~\ref{tab:model_ids}. All models were called through the organization’s official APIs (except for Llama, for which we used the Together API).

\paragraph{Hyperparameter Details.} In all generation tasks, the temperature was set to the default value in the organization's API. For response generation and context generation, we sampled a maximum of 2048 tokens while for model evaluation, we sampled a maximum of 512 tokens.

\paragraph{Prompts.} The prompt used for classifying queries into query types is provided in Table~\ref{tab:query_type_classification_prompt}. The prompt for generating follow-up QAs for a query is provided in Table~\ref{tab:contextual_followup_qa_prompt}. For getting autorater judgements without and with context, the prompts are provided in Tables~\ref{tab:eval_judgement_wo_ctx_prompt} and \ref{tab:eval_judgement_w_ctx_prompt} respectively. Finally, the prompt to compute the number of constraints from the context met by a response is provided in Table~\ref{tab:criteria_satisfaction_judgement_prompt}. For the analyses presented in sections~\ref{sec:default} and \ref{sec:sensitivity}, we provide the prompt used to filter queries in Table~\ref{tab:query_followup_evaluation_prompt}, and the prompt used to rate responses on a Likert scale is in Table~\ref{tab:response_quality_rating_prompt}.

\begin{table}[ht!]
\centering
\scalebox{.75}{
   \setlength\extrarowheight{-2pt}
   \setlength\tabcolsep{5pt} 
\begin{tabular}{ll}
\rowcolor{gray!40}
    \toprule
    \textbf{Model Name} & \textbf{Identifier} \\
    \midrule
        \texttt{GPT-4o} & \raggedright\arraybackslash\ttfamily gpt-4o-2024-05-13 \\ 
        \texttt{Gemini-1.5-Flash} & \raggedright\arraybackslash\ttfamily gemini-1.5-flash-exp-0827 \\
        \texttt{Claude-3.5-Sonnet} & \raggedright\arraybackslash\ttfamily claude-3.5-sonnet \\
        \texttt{Llama-3.1-405B} & \raggedright\arraybackslash\ttfamily Meta-Llama-3.1-405B-Instruct-Turbo \\
        \texttt{Gemma-2-27B} & \raggedright\arraybackslash\ttfamily gemma-2-27b-it \\
        \texttt{Jamba-1.5-Large} & \raggedright\arraybackslash\ttfamily jamba-1.5-large \\
        \texttt{Qwen2-27B-Instruct} & \raggedright\arraybackslash\ttfamily Qwen2-72B-Instruct \\
    \bottomrule
\end{tabular}
}
\caption{List of models used in our experiments and their official identifiers.}
\label{tab:model_ids}
\end{table}

\setlength{\tabcolsep}{6pt}
\renewcommand{\arraystretch}{1}
\definecolor{Lavender}{rgb}{0.9,0.9,0.98}

\begin{table*}[t!]
\centering
\scalebox{0.75}{
\rowcolors{2}{Lavender!55}{white}
\begin{tabular}{p{6cm} p{14cm}}
\rowcolor{Lavender!80}
\multicolumn{1}{l}{\textbf{Contextual Attribute}} & \multicolumn{1}{l}{\textbf{Follow-Up QA}} \\ \toprule
Level of Detail & Q: How much detail do you prefer in the response? \newline A: ["One-sentence answer", "Key points only", "Moderate detailed", "Extensive detail"] \\ \hline
User Expertise & Q: What is your level of expertise on this topic? \newline A: ["Complete beginner", "Basic understanding", "Intermediate", "Advanced", "Expert"] \\ \hline
Length & Q: What is your preferred length for the response? \newline A: ["One sentence", "2-3 sentences", "One paragraph (>3 sentences)", "Several paragraphs"] \\ \hline
Format of response & Q: What format would you prefer the response to be in? \newline A: ["Bulleted list", "Numbered steps", "Paragraph text", "Table or chart"] \\ \hline
Style & Q: What style of response do you prefer? \newline A: ["Formal", "Informal", "Conversational", "Academic", "Technical"] \\ \hline
Intended Audience & Q: Who is the intended audience for this response? \newline A: ["General public", "Children", "Students", "Professionals / Experts"] \\ \hline
Geographical / Regional Context & Q: What region or country should this response be based on? \newline A: ["North America", "Europe", "Asia", "Africa", "Latin America"] \\ \hline
Cultural Context & Q: What cultural perspective should be considered in the response? \newline A: ["Western culture", "Eastern culture", "Indigenous culture", "Multicultural perspective"] \\ \hline
Age Group & Q: Which age group should this response be relevant for? \newline A: ["Children", "Teenagers", "Young adults", "Middle-aged adults", "Seniors"] \\ \hline
Economic Context & Q: What economic situation should this response be relevant for? \newline A: ["Low-income", "Middle-income", "High-income", "Budget-conscious"] \\ \hline
Political Context & Q: What political context should this response consider? \newline A: ["Liberal", "Conservative", "Centrist", "Socialist"] \\ \hline
Gender & Q: Should the response consider any specific gender perspective? \newline A: ["Male", "Female", "Non-binary", "Gender-neutral"] \\ \bottomrule
\end{tabular}
}
\caption{Contextual attributes and corresponding follow-up QA considered for the implicit context analysis presented in Sections~\ref{sec:default} and \ref{sec:sensitivity}.}
\label{tab:default_context_qa}
\end{table*}


\begin{table*}[ht!]
    \centering
    \footnotesize
    \rowcolors{2}{gray!15}{gray!15}
    \begin{tabular}{p{2\columnwidth}}
        \toprule
        \textbf{Prompt for Classifying Query Types} \\
        \midrule
        \raggedright \texttt{You will be shown a query issued by a real user to a language model. You need to answer what query type(s) this query belongs to, from the list below. \linebreak \linebreak - Ambiguous: Queries which can be interpreted in different ways, that cause confusion about what is being asked. \linebreak - Incomplete: Queries which lack information that is essential to understand the intent of the query. Note these are different from ambiguous queries, which need clarification due to multiple possible interpretations. \linebreak - Subjective: Queries whose responses can be influenced by personal beliefs and perspectives. \linebreak - Open-ended: Queries which require detailed responses and lack a single, concise answer. \linebreak - Closed-ended: Queries which require an unambiguous and concise answer. \linebreak \linebreak Note that a single query can belong to multiple query types. Provide your output as a list with the query types that the query belongs to.} \linebreak \linebreak
        \#\#\#\linebreak
        \linebreak \texttt{Query: best team in the league \linebreak Query Types: ["Incomplete", "Subjective", "Closed-ended"]} \linebreak
        \texttt{Query: [QUERY] \linebreak Query Types:}
    \end{tabular}
    \caption{Prompt for classifying queries into different query types.}
    \label{tab:query_type_classification_prompt}
\end{table*}

\begin{table*}[ht!]
    \centering
    \footnotesize
    \rowcolors{2}{gray!15}{gray!15}
    \begin{tabular}{p{2\columnwidth}}
        \toprule
        \textbf{Prompt for Generating Follow-up QAs} \\
        \midrule
        \raggedright \texttt{You will be shown a query issued by a real user to a language model. Imagine that you are required to answer this query. First, you need to answer whether it would be helpful to know context surrounding this query to give a useful response. The context can be about the user (eg, their background, age, language fluency, location, profession, expertise etc), their intent / preferences for the response (eg, query intent, text formatting/style, structure, length, presence of citations, or any other open-ended criteria) or information missing that is required to respond to a query or resolve ambiguity in the query. Queries that are objective, closed-ended or have straightforward answers should not require context. \linebreak \linebreak Answer in Yes or No for whether context is required and generate context if the answer is Yes. This context should be formatted as follow-up question answer pairs, where you ask the most important questions first and list plausible answers to these questions. \linebreak \linebreak Here are criteria that individual questions need to satisfy:\linebreak - salient: The question should ask about information that would be useful to adapt the query's response to the user's needs and background.\linebreak - influential: The answer to this question should directly influence the response. With different answers to this question, the response to the query would need to be phrased differently.\linebreak \linebreak Here are the criteria that the list of questions needs to satisfy:\linebreak - sufficient: There should be enough important questions to cover a large space of possible contexts for the query.\linebreak - ranked in order of salience: the questions should be ranked in the order of their importance.\linebreak \linebreak Here are the criteria that each answer set needs to satisfy:\linebreak - plausible answers: The answer set should represent a realistic set of answers to the question, such that a real user would answer the question with any of the choices. Do not generate answer choices such as "Other" which are uninformative.\linebreak - discrete answer space: The possible answers to the question should be discrete, short strings.\linebreak - diverse coverage: The answer set should be a representative set of possible answers to the question, such that each answer choice would elicit different responses to the original query.\linebreak \linebreak Generate up to 10 follow-up QA pairs and they should all meet the above criteria. Each QA pair should be such that it is easy to check whether the QA is incorporated in a candidate response.} \linebreak \linebreak
        \textbf{Example Follow-up QA:} \linebreak
        \linebreak \texttt{Query: best team in the football league \linebreak Need for Context: Yes \linebreak Context: Q: Which league are you referring to? A: ["English Premier League", "La Liga", "Bundesliga", "Italian Serie A", "MLS", "UEFA"] \linebreak Q: How do you define "best"? A: ["Most recent wins", "Number of championships won", "Goal difference", "Squad strength"] \linebreak Q: Do you want the best team based on current form or overall historical performance? A: ["Current form", "Historical performance"] \linebreak Q: Are you asking about men’s football or women’s football? A: ["Men’s football", "Women’s football"] ... \linebreak \linebreak Query: How do antibiotics work against bacteria? \linebreak Need for Context: Yes \linebreak Context: Q: What is your background in biology or medicine? A: ["No background", "High school level", "College level", "Medical or professional background"] \linebreak Q: What is your purpose for asking this question? A: ["For a class", "Personal knowledge", "Professional/medical use", "To explain to someone else"] \linebreak Q: What level of detail are you looking for in the explanation? A: ["Basic overview", "Intermediate (some scientific terms)", "Detailed (in-depth biological mechanisms)"] ... } \linebreak
    \end{tabular}
    \caption{Prompt for generating contextual follow-up questions for user queries.}
    \label{tab:contextual_followup_qa_prompt}
\end{table*}

\begin{table*}[ht!]
    \centering
    \footnotesize
    \rowcolors{2}{gray!15}{gray!15}
    \begin{tabular}{p{2\columnwidth}}
        \toprule
        \textbf{Prompt for Autorater Preference Judgements (setting \texttt{NoCtxGen-NoCtxEval})} \\
        \midrule
        \raggedright \texttt{You will be given a query issued by a real user to a language model. You will also be given two model responses to this query, and you will need to judge which response is better. \linebreak \linebreak IMPORTANT: You should produce the final judgement as a dictionary in precisely this format (with **): "**output: \{"judgement": "\_" \}**", where you should fill in the spaces with either "Response 1" if Response 1 is better, "Response 2" if Response 2 is better or "Tie" if both responses are equally good or equally bad. Only the three choices "Response 1", "Response 2" and "Tie" are valid. Make note of the \textbf{**} required to enclose the output dictionary. After generating the output, provide a brief justification of your judgement.} \linebreak
        \linebreak \texttt{Query: [QUERY] \linebreak Response 1: [RESPONSE 1] \linebreak Response 2: [RESPONSE 2] \linebreak Judgement: **output: \{"judgement": "\_" \}** \linebreak Justification: [JUSTIFICATION]} \linebreak
    \end{tabular}
    \caption{Prompt for getting autorater preference judgements for the setting \texttt{NoCtxGen-NoCtxEval}.}
    \label{tab:eval_judgement_wo_ctx_prompt}
\end{table*}

\begin{table*}[ht!]
    \centering
    \footnotesize
    \rowcolors{2}{gray!15}{gray!15}
    \begin{tabular}{p{2\columnwidth}}
        \toprule
        \textbf{Prompt for Autorater Preference Judgements (settings \texttt{NoCtxGen-CtxEval} and \texttt{CtxGen-CtxEval})} \\
        \midrule
        \raggedright \texttt{You will be given a query issued by a real user to a language model and the context under which the query was issued. This context will be presented in the form of follow-up questions and the user's answers to these questions. The context provides information about the user's intent, preferences and background. \linebreak \linebreak You will be given two model responses to this query, and you will need to judge which response more accurately and completely incorporates the information from the query and context. To evaluate the responses, you should first check whether the answer to each of the follow-up questions in the context is incorporated well in each response. Then, you should choose the response which incorporates more of the constraints from the context and provides the most relevant and complete answer to the query. \linebreak \linebreak IMPORTANT: You should produce the final judgement as a dictionary in precisely this format (with **): "**output: \{"judgement": "\_" \}**", where you should fill in the spaces with 1) "Response 1" if Response 1 is better, 2) "Response 2" if Response 2 is better or 3) "Tie" if both responses are equally good or equally bad. Only the three choices "Response 1", "Response 2" and "Tie" are valid. Make note of the \textbf{**} required to enclose the output dictionary. After generating the output, provide a brief justification of your judgement that mentions which aspects of the context were better incorporated by the chosen response, or why the responses are equally good or equally lacking.} \linebreak
        \linebreak \texttt{Query: [QUERY] \linebreak Context: [CONTEXT] \linebreak Response 1: [RESPONSE 1] \linebreak Response 2: [RESPONSE 2] \linebreak Judgement: **output: \{"judgement": "\_" \}** \linebreak Justification: [JUSTIFICATION]} \linebreak
    \end{tabular}
    \caption{Prompt for getting autorater preference judgements for the context-aware evaluation settings (\texttt{NoCtxGen-CtxEval} and \texttt{CtxGen-CtxEval}).}
    \label{tab:eval_judgement_w_ctx_prompt}
\end{table*}

\begin{table*}[ht!]
    \centering
    \footnotesize
    \rowcolors{2}{gray!15}{gray!15}
    \begin{tabular}{p{2\columnwidth}}
        \toprule
        \textbf{Prompt for computing the number of constraints in a context that are met by a response} \\
        \midrule
        \raggedright \texttt{You will be given a query issued by a real user and the context under which the query was issued. This context will be presented in the form of follow-up questions and the user's answers to them. \linebreak \linebreak You will be given a model response to this query, and you will need to judge how many of the criteria in the follow-up questions are addressed by the response. So if the response incorporates 5 of the follow-up questions completely, you should output 5. If it incorporates 2 of the follow-up questions, you should output a 2. If it does not address any of the follow-up questions, you should rate it as a 0. \linebreak \linebreak IMPORTANT: You should first generate a single number, which is the total number of constraints satisfied. After generating this number, provide a very brief justification for your answer.} \linebreak
        \linebreak \texttt{Query: [QUERY] \linebreak Context: [CONTEXT] \linebreak Response: [RESPONSE] \linebreak Output:}\linebreak
    \end{tabular}
    \caption{Prompt for evaluating how many follow-up QAs in the context are satisfied by a response.}
    \label{tab:criteria_satisfaction_judgement_prompt}
\end{table*}

\begin{table*}[ht!]
    \centering
    \footnotesize
    \rowcolors{2}{gray!15}{gray!15}
    \begin{tabular}{p{2\columnwidth}}
        \toprule
        \textbf{Prompt for checking the importance of a contextual attribute for a query} \\
        \midrule
        \raggedright \texttt{You will be given a query from a real user to a language model, along with a follow-up question that can be asked to the user. The follow-up question will have a set of answer choices. Your task is to answer the following three questions: \linebreak \linebreak 1) Is it important to know the user's answer to the follow-up question to provide a useful response to the original query? \linebreak 2) Is the query independent of the answer choices? If the query already implies a specific answer choice, it is not independent. \linebreak 3) Is the query well-formed? A well-formed query clearly expresses an information need, even if it is not fully fluent, unambiguous, or fully specified. Queries not in English are not considered well-formed. \linebreak \linebreak IMPORTANT: Please provide the final output in the following dictionary format: \{"1": "Yes/No", "2": "Yes/No", "3": "Yes/No"\}}. \linebreak
        \linebreak \texttt{Query: [QUERY] \linebreak Follow-up Question: [QUESTION] \linebreak Output:} \linebreak
    \end{tabular}
    \caption{Prompt for evaluating the importance, independence, and well-formedness of queries with follow-up questions.}
    \label{tab:query_followup_evaluation_prompt}
\end{table*}

\begin{table*}[ht!]
    \centering
    \footnotesize
    \rowcolors{2}{gray!15}{gray!15}
    \begin{tabular}{p{2\columnwidth}}
        \toprule
        \textbf{Prompt for rating response relevance based on query and context} \\
        \midrule
        \raggedright \texttt{You will be given a query issued by a real user to a language model and the context under which the query may have been issued. This context will be presented in the form of a follow-up question issued to the user and possible answers to this question. \linebreak \linebreak You will be given a model response to this query, and you will need to judge the quality of this response corresponding to each follow-up question-answer pair. Rate the response on a scale of 1-5 on the following axis: \linebreak \linebreak * Relevance: How relevant is the response to addressing the query and context? \linebreak \hspace*{0.5cm} * 1: The response is not helpful in responding to the query and context at all. \linebreak \hspace*{0.5cm} * 2: The response provides limited help, missing important information from the query or context. \linebreak \hspace*{0.5cm} * 3: The response is somewhat helpful, offering useful information but lacking thoroughness or depth for the query and context. \linebreak \hspace*{0.5cm} * 4: The response is helpful, addressing most of the query and context adequately. \linebreak \hspace*{0.5cm} * 5: The response is highly helpful, fully addressing the query and context with thorough and useful information. \linebreak \linebreak IMPORTANT: You should produce the final output as a dictionary in precisely this format (with **): \texttt{[OUTPUT\_FORMAT]}, where you should fill in the spaces with ratings for each one of the possible answers to the follow-up question. Make note of the ** required to enclose the output dictionary.} \linebreak
        \linebreak \texttt{Query: [QUERY] \linebreak Context: [CONTEXT] \linebreak Response: [RESPONSE] \linebreak Judgement:} \linebreak
    \end{tabular}
    \caption{Prompt for rating response relevance based on query and context.}
    \label{tab:response_quality_rating_prompt}
\end{table*}

\end{document}